\documentclass[twocolumn]{theme/bioRxiv}

\usepackage{multirow} %

\usepackage{amsmath,amsfonts,bm}

\def\eqref#1{equation~\ref{#1}}

\def\1{\bm{1}}

\DeclareMathAlphabet{\mathsfit}{\encodingdefault}{\sfdefault}{m}{sl}
\SetMathAlphabet{\mathsfit}{bold}{\encodingdefault}{\sfdefault}{bx}{n}

\usepackage{url}
\usepackage{subcaption}
\usepackage{booktabs}
\usepackage{algorithm}
\usepackage{algpseudocode}
\usepackage{tabularx}
\usepackage{enumitem}
\usepackage{adjustbox} %

\usepackage{tikz}
\usetikzlibrary{
    positioning,
    calc,
    arrows.meta,
    shapes.geometric,
    decorations.pathreplacing,
    backgrounds,
    fit
}

\definecolor{abcblue}{RGB}{52, 101, 164}
\definecolor{abcorange}{RGB}{245, 121, 0}
\definecolor{abcgreen}{RGB}{78, 154, 6}
\definecolor{tabaccent}{RGB}{100, 100, 100}

\newcommand{\msmc}{\textsc{M-SMC}}
\newcommand{\dsmc}{\textsc{D-SMC}}
\newcommand{\smcabc}{\textsc{SMC-ABC}}
\newcommand{\olb}{\textsc{OLB-300}}
\newcommand{\sd}{\textsc{SD}}
\newcommand{\multicare}{\textsc{MultiCare}}
\newcommand{\abi}{\textsc{ABI}}
\newcommand{\best}[1]{\textbf{#1}}

\titleformat{\section}
  {\Large\sffamily\bfseries}
  {\thesection}
  {0.6em}
  {#1}
  []
\renewcommand{\thesubsection}{\thesection.\arabic{subsection}}
\titleformat{\subsection}
  {\sffamily\bfseries}
  {\thesubsection}
  {0.5em}
  {#1}
  []
\def\subsectionmark#1{\relax} %

\makeatletter
\def\aBioRXiv{{\footnotesize
    \footerfont \@leadauthor\ifnum \value{authors} > 2\hspace{2pt}\textit{et al.}\fi\hspace{7pt}|\hspace{7pt}\today\hspace{7pt}|\hspace{7pt}\thepage\textendash\pageref{LastPage}
  }}
\def\bBioRXiv{{\footnotesize
    \footerfont \thepage
  }}
\def\cBioRXiv{{\footnotesize
    \footerfont \thepage
  }}
\makeatother

\hypersetup{
  pdftitle={Uncertainty-Aware Calibrated Clinical Text Classification with Large Language Models},
  pdfauthor={Mridul Sharma, Adeetya Patel, Zaneta D'souza, Samira Abbasgholizadeh Rahimi, Siva Reddy, Sreenath Madathil},
  pdfkeywords={uncertainty quantification, large language models, approximate Bayesian computation, sequential Monte Carlo, clinical text classification, calibration}
}

\begin{document}

\leadauthor{Sharma}

\title{Uncertainty-Aware Calibrated Clinical Text Classification with Large Language Models}
\shorttitle{Uncertainty-aware clinical text classification with LLMs}

\author[1,*]{Mridul Sharma}
\author[1]{Adeetya Patel}
\author[1]{Zaneta D'souza}
\author[1,3]{Samira Abbasgholizadeh Rahimi}
\author[2,3]{Siva Reddy}
\author[1]{Sreenath Madathil}
\affil[1]{Faculty of Dental Medicine and Oral Health Sciences, McGill University, Montreal, Canada}
\affil[2]{School of Computer Science, McGill University, Montreal, Canada}
\affil[3]{Mila--Quebec Artificial Intelligence Institute, Montreal, Canada}
\date{}

\maketitle

\begin{abstract}
Large language models are increasingly used for clinical text classification, where overconfident misclassifications can directly affect patient care. Existing black-box uncertainty methods attach a confidence score to a fixed LLM prediction using softmax probabilities, verbalised confidence, prompt agreement, or generation consistency. These signals are often poorly calibrated and offer no mechanism for combining model evidence with prior clinical belief. We instead formulate closed-set clinical classification as likelihood-free posterior inference over diagnostic hypotheses. A prompt-conditioned LLM is treated as a class-conditional stochastic simulator: for each candidate diagnosis it generates synthetic clinical descriptions, which are compared with the observed case in an embedding summary space. Sequential Monte Carlo Approximate Bayesian Computation then yields a posterior over diagnoses from which both the prediction and its uncertainty are derived. We instantiate three variants spanning a cost--fidelity spectrum: multinomial SMC (\msmc{}), Dirichlet SMC (\dsmc{}), and an amortised Dirichlet posterior network (\abi{}) for deployment-time inference. Across three clinical benchmarks and six open LLMs, the inferred posterior separates correct predictions from errors more sharply than black-box confidence, with misclassification-detection AUROC on \multicare{} of $0.923$ for \msmc{} against $0.760$ for softmax and $0.584$ for a consistency baseline. The posterior also accepts an explicit clinician prior, recovering the correct diagnosis from a deliberately misleading one, and remains diffuse on out-of-distribution presentations. We release \olb{}, an expert-validated clinical uncertainty-quantification benchmark of 300 oral-lesion vignettes.
\end{abstract}


\begin{corrauthor}
mridul.sharma\at mail.mcgill.ca
\end{corrauthor}

\section{Introduction}
Clinical decision support with large language models (LLMs) reduces, in
many high-value settings, to closed-set text classification: mapping a
free-text symptom description or case narrative to one of a known set of
diagnostic categories. Unlike open-ended generation, the hypothesis space
is finite, but a predicted label alone is insufficient. Downstream
clinical decisions such as referral thresholds, biopsy recommendations,
and triage priority depend on the magnitude of risk. A posterior that
assigns $p(\text{cancer}) = 0.91$ implies a categorically different
action than one that assigns $p(\text{cancer}) = 0.54$, even when both
produce the same top prediction. Calibrated probabilities over the full
label set are therefore an operational requirement, not a modelling
convenience.

Calibrated point probabilities are necessary but not sufficient. For
example, in a diagnostic setting a scalar
$p(\text{disease} \mid \text{symptom}_{\text{obs}})$ collapses two
clinically distinct situations: irreducible diagnostic ambiguity, where
symptoms genuinely overlap across plausible diagnoses and the appropriate
response is further testing; and model unreliability, where the input is
atypical or out-of-distribution and the appropriate response is deferral.
Second-order uncertainty around the class probability vector
distinguishes these by decomposing predictive uncertainty into aleatoric
(class overlap) and epistemic (estimate reliability) components.

Current methods do not provide this object. Score-based
\citep{guo2017calibration,zhang2024calibrating}, elicitation-based
\citep{tian2023just,xiong2024can} and consistency-based approaches
\citep{kuhn2023semantic,manakul2023selfcheckgpt,xiao2025consistency}
all attach a confidence value to a fixed prediction (\S\ref{sec:related}).
The consistency family is the most instructive: in closed-set
classification it reduces to the entropy of the empirical label
distribution under stochastic decoding, a quantity derived from the
model's outputs under a fixed prompt rather than from any comparison of
competing hypotheses against the observed case. A model can therefore be
perfectly consistent and systematically wrong whenever all its
generations share a prompt bias or a missing piece of evidence. None of
these methods perform inference over diagnostic hypotheses, none admit a
prior belief, and none produce uncertainty coupled to hypothesis
competition.

The formulation these methods approximate, but do not implement, is
posterior inference over the label set: given $x_{\text{obs}}$, compute
$p(y \mid x_{\text{obs}})$ over the finite diagnostic set $\mathcal{Y}$,
combining observed evidence with prior belief. One of the barriers is
likelihood intractability, since the probability that a prompted LLM
would generate a particular observed string under a candidate diagnosis
is not available in closed form. We address this with Approximate
Bayesian Computation~\citep{beaumont2009adaptive}, the standard framework
for likelihood-free inference when simulation is tractable but likelihood
evaluation is not. The enabling observation is that a prompted LLM is a
natural class-conditional stochastic simulator: for any label
$y \in \mathcal{Y}$ it can generate synthetic descriptions
$x_{\text{sim}} \sim p_{\text{LLM}}(\cdot \mid y)$, even though it may not
be able to score an arbitrary observed string under that hypothesis.
Those simulations are compared with the observed case in a sentence
embedding space via cosine distance, and SMC-ABC then refines a particle
population over the label set, concentrating posterior mass on classes
whose simulations best match the observation
(Figure~\ref{fig:schematic}). The result is a posterior whose
calibration can be evaluated empirically, rather than a confidence score
attached to an unchanged model prediction.

Because the output is a posterior rather than a score, three things
follow that existing methods cannot offer: entropy that reflects genuine
competition among hypotheses and therefore supports abstention, an
explicit prior that lets clinician knowledge be combined with model
evidence, and (in the Dirichlet variants) a distribution over the
simplex that stays diffuse outside the modelled label space.

\paragraph{Contributions}
\begin{enumerate}\setlength{\itemsep}{1pt}\setlength{\topsep}{1pt}
    \item We formulate closed-set clinical text classification as likelihood-free posterior inference over the diagnostic label set, treating the prompted LLM as a class-conditional simulator and applying SMC-ABC to obtain a posterior without access to model internals. This turns uncertainty from a score attached to a fixed prediction into a property of the inference itself, and admits an explicit clinical prior.
    \item We instantiate the framework at three points on a cost--fidelity spectrum: categorical (\msmc{}), Dirichlet (\dsmc{}), and amortised (\abi{}). Across three benchmarks and six open LLMs, the resulting posterior separates correct predictions from errors more sharply than black-box confidence (misclassification-detection AUROC $0.923$ against $0.760$ for softmax, a gap that persists under accuracy-controlled excess-AURC), recovers the true diagnosis from a misleading prior, and reaches maximal entropy on out-of-distribution inputs, at $18\,\mu$s per case under \abi{}.
    \item We release \olb{}, an expert-validated clinical uncertainty-quantification benchmark of 300 oral-lesion vignettes, reviewed blind by an oral pathologist on completeness and medical plausibility.
\end{enumerate}

\section{Background and Related Work}
\label{sec:related}
\paragraph{Uncertainty quantification for LLMs.}
Score-based methods use logits or softmax outputs as confidence estimates
\citep{guo2017calibration,desai2020calibration}, signals that can be
degraded by RLHF alignment \citep{zhang2024calibrating}. Post-hoc
calibration methods such as Platt scaling and temperature scaling rescale a
fixed predictor \citep{platt1999probabilistic,guo2017calibration}, but
cannot by themselves recover errors from missing evidence or semantic
ambiguity. Elicitation-based methods prompt the model to verbalize
confidence \citep{tian2023just,xiong2024can}, which can overestimate
accuracy in medical question answering \citep{savage2024large}.
\paragraph{Black-box sampling and consistency methods.}
A third family estimates uncertainty from variability across
generations, prompts, or perturbations: prompt
consistency~\citep{zhou2022prompt}, semantic
entropy~\citep{kuhn2023semantic}, SelfCheck-GPT-style hallucination
detection~\citep{manakul2023selfcheckgpt}, perturbation-based
UQ~\citep{gao2024spuq}, and similarity-based
aggregation~\citep{lin2024generating,bhattacharjya2024consistency,bhattacharjya2025simba,xiao2025consistency}.
In closed-set classification, semantic equivalence reduces to the
predicted class label, so semantic-entropy style uncertainty can be
approximated by the entropy of the empirical label distribution under
repeated stochastic decoding. We include this as the Elicited Fidelity
(EF) baseline~\citep{zhang2024calibrating}. Our method differs by
performing prior-conditioned posterior inference over diagnostic
hypotheses rather than aggregating samples from a fixed prompt.
\paragraph{Uncertainty quantification for encoder-based clinical
classifiers.}
A parallel line of work quantifies uncertainty for supervised encoder
models on clinical text, combining probability-based scores, MC dropout,
density-based scores, and hybrid aleatoric--epistemic estimators for
selective prediction and abstention~\citep{vazhentsev2026uncertainty}.
These occupy a different point on the data-access/compute trade-off:
encoder methods need labelled task data to fine-tune on but are cheap at
inference, whereas \msmc{}/\dsmc{} need no task training and pay per
case in simulation. Where abundant labelled data for the exact
diagnostic panel exist, a fine-tuned encoder with post-hoc calibration
is a strong and cheaper choice; \S\ref{sec:embedding-ablation} raises
the same caveat for calibrated embedding classifiers.

\paragraph{Approximate Bayesian Computation (ABC).}
ABC is a likelihood-free inference framework for settings in which
simulation is possible but likelihood evaluation is
unavailable~\citep{beaumont2009adaptive}. SMC-ABC
refines a particle population by progressively tightening acceptance
thresholds. ABC has been applied in population genetics, epidemiology,
and astrophysics, and recent simulation-based inference work extends
this paradigm with neural density
estimators~\citep{deistler2025simulation}.
Closest to our setting, \citet{shen2023reliable} use SMC-ABC for
likelihood-free prompt tuning, where the unknowns are soft prompt
embeddings. Our objective is different: we treat the prompted LLM as a
fixed simulator and infer a posterior over diagnostic class hypotheses,
so the quantity being inferred is a distribution over labels rather than
over prompts. A central design choice in
ABC is the summary statistic used to compare simulated and observed
data~\citep{fearnhead2012constructing}; in our setting the sentence
embedding serves this role, so the choice of embedding model is a
modelling decision analogous to summary-statistic selection in classical
ABC (Appendix~\ref{app:embeddings}).

\section{Method}
\label{sec:method}
\subsection{Re-framing Text Classification as Simulation-Based
Inference}

Let $\mathcal{Y} = \{1,\dots,K\}$ denote the diagnostic label set and
$x_{\mathrm{obs}}\in\mathcal{X}$ an observed clinical description. We
treat the prompted LLM as a class-conditional stochastic simulator:
given a hypothesised label $y\in\mathcal{Y}$ it generates
$x_{\mathrm{sim}}\sim p_{\mathrm{LLM}}(\cdot\,|\,y)$. Because the
likelihood $p_{\mathrm{LLM}}(x_{\mathrm{obs}}\,|\,y)$ is not available in
closed form, inference proceeds through simulation.

We compare observed and simulated descriptions through an embedding
$\phi:\mathcal{X}\to\mathbb{R}^d$. With
$z_{\mathrm{obs}}=\phi(x_{\mathrm{obs}})$ and
$z_{\mathrm{sim}}=\phi(x_{\mathrm{sim}})$, define the discrepancy
$\rho(z_{\mathrm{sim}},z_{\mathrm{obs}})=1-\cos(z_{\mathrm{sim}},z_{\mathrm{obs}})$.
Given a prior $\pi(y)$ over diagnoses and an ABC tolerance $\epsilon$, the
resulting ABC target is
\begin{equation}
\begin{split}
p_\epsilon(y\,|\,x_{\mathrm{obs}})
\propto{}&
\pi(y)\,
\mathbb{E}_{x_{\mathrm{sim}}\sim p_{\mathrm{LLM}}(\cdot\,|\,y)}
\Bigl[\\
&\quad
\mathbf{1}\bigl\{
\rho(\phi(x_{\mathrm{sim}}),\phi(x_{\mathrm{obs}}))\le\epsilon
\bigr\}
\Bigr].
\end{split}
\label{eq:abc-target}
\end{equation}
Posterior mass accrues to hypotheses whose simulations lie close to the
observation in embedding space, and concentrates as $\epsilon$ decreases.
The result is posterior inference under the prompt-conditioned simulator
and embedding summary, not under the true clinical data-generating
process, a distinction we return to in \S\ref{sec:limitations}. This
single object supplies both the prediction (via a decision rule) and the
uncertainty (via the mass on competing diagnoses), which makes the
framework a classifier with inherent UQ rather than a confidence score
attached to an unchanged LLM output.

\subsection{Posterior representations and amortised inference}

\paragraph{Multinomial \msmc{}.}
Each particle is a discrete class index $y\in\mathcal{Y}$, drawn initially
from the prior $\pi(y)$. As with \dsmc{} below, this prior may be
non-informative or may carry clinician-supplied beliefs, and
\S\ref{sec:prior} evaluates both cases. Each particle is scored by
drawing a fresh simulation from its own hypothesised class and comparing
that simulation with the observation; \msmc{} therefore does not use the
class-mean construction introduced for \dsmc{} below. After the final
SMC iteration the weighted particle population defines an empirical
categorical posterior,
\[
\hat{p}(y\,|\,x_{\mathrm{obs}})
=
\sum_{s=1}^{S} w_{T,s}\,\mathbf{1}\{\tilde y_{T,s}=y\},
\]
where $T$ is the index of the final SMC iteration, $s\in\{1,\dots,S\}$
indexes the $S$ particles of that final population, $\tilde y_{T,s}$ is
the class label carried by particle $s$ at iteration $T$, and $w_{T,s}$
its normalised weight ($\sum_s w_{T,s}=1$). This representation is
directly interpretable as posterior mass over diagnostic labels and can
capture multimodality when several diagnoses remain plausible.

\paragraph{Dirichlet \dsmc{}.}
Each particle is a probability vector
$\theta\in\Delta^{|\mathcal{Y}|-1}$, with initial particles sampled from
a Dirichlet prior $\theta\sim\mathrm{Dir}(\alpha)$. Here, $\alpha$ may be
chosen to encode a non-informative prior or clinician-supplied prior
beliefs over diagnoses. For a particle $\theta$, we construct a
simplex-weighted representation
\begin{gather*}
\tilde z(\theta)=\sum_{k=1}^{K}\theta_k\,\bar z_k,\\
\bar z_k=\frac{1}{M}\sum_{m=1}^{M}\phi\!\left(x^{(m)}_k\right),
\quad
x^{(m)}_k\sim p_{\mathrm{LLM}}(\cdot\mid k),
\end{gather*}
so that $\bar z_k$ is simply the mean embedding of the $M$ descriptions
the simulator generates for class $k$, and $\tilde z(\theta)$ is a
$\theta$-weighted combination of those $K$ class means. We call $\bar z_k$
a class-conditional centroid only in this descriptive sense; it is
estimated entirely from \emph{simulated} text, and the observation enters
nowhere in its construction. The observation is used only once, in the
discrepancy $\rho(\tilde z(\theta), z_{\mathrm{obs}})$ on which acceptance
is based. and the SMC procedure
produces a weighted particle approximation to a posterior over $\theta$.
We report the posterior mean for prediction, while the spread of the
weighted particles provides a second-order summary of uncertainty over
class probabilities (Appendix~\ref{sec:second-order}). This smoother
simplex-level representation is useful for analysing ambiguity and
out-of-distribution behaviour, at the cost of greater sensitivity to the
prior (\S\ref{sec:prior}).

\paragraph{Amortised Bayesian inference (\abi{}).}
Both \msmc{} and \dsmc{} require a separate SMC-ABC run for each test
case. This is appropriate for high-fidelity inference, but expensive for
deployment settings where many cases must be processed quickly. We
therefore include an amortised variant, \abi{}, which learns a direct map
from the observed embedding to a Dirichlet posterior over the simplex.

Specifically, \abi{} approximates the simulator-induced posterior with
the variational family
\begin{equation}
\begin{gathered}
q_\varphi(\theta \mid z_{\mathrm{obs}})
=
\mathrm{Dir}\!\left(\theta;\,\bm{\alpha}_\varphi(z_{\mathrm{obs}})\right),\\
\bm{\alpha}_\varphi(z_{\mathrm{obs}}) \in \mathbb{R}_{+}^{K},
\end{gathered}
\label{eq:abi}
\end{equation}
where $z_{\mathrm{obs}}=\phi(x_{\mathrm{obs}})$ and
$\bm{\alpha}_{\varphi}$ is a small multilayer perceptron with a
softplus output layer.

It is worth being explicit about what is being amortised, since two
distinct Bayesian objects are in play. The target is the posterior
produced by the ABC procedure itself: we run \dsmc{} on a training
partition, fit a Dirichlet to each resulting particle posterior, and
train $\bm{\alpha}_\varphi$ to regress from $z_{\mathrm{obs}}$ to those
Dirichlet parameters (Appendix~\ref{sec:abi-details}). \abi{} is
therefore an amortised approximation \emph{of an ABC posterior}, and it
inherits whatever that posterior encodes, including the prior used during
the \dsmc{} runs. The network itself is deterministic: $\varphi$ is a
point estimate obtained by maximum likelihood, and we place no
distribution over the weights. The Bayesian content lives in the target
$q_\varphi(\theta\mid z_{\mathrm{obs}})$ over the simplex, not in the
network that predicts it. One practical consequence is that the prior is
fixed at training time, which is why test-time prior injection is
available in \msmc{} and \dsmc{} but not in \abi{}
(Table~\ref{tab:variants}).

\begin{figure*}[t]
\centering
\begin{tikzpicture}[
  font=\scriptsize,
  >={Stealth[length=1.5mm,width=1.2mm]},
  ar/.style={->, line width=0.5pt, draw=black!65},
  stage/.style={draw=black!15, fill=black!2, rounded corners=3pt, line width=0.6pt},
  card/.style={draw=black!35, fill=white, rounded corners=1.8pt, line width=0.5pt,
               align=center, inner sep=2pt},
  hyp/.style={draw=abcblue!60, fill=abcblue!8, rounded corners=1.8pt, line width=0.5pt,
              align=center, inner sep=1.5pt, minimum width=12mm, minimum height=3.9mm},
  llm/.style={draw=abcorange!70, fill=abcorange!12, rounded corners=1.8pt, line width=0.6pt,
              align=center, inner sep=2.5pt},
  tag/.style={font=\scriptsize\bfseries, anchor=north west, text=black!75},
  sub/.style={draw=black!25, fill=white, rounded corners=2pt, line width=0.5pt},
]

\begin{scope}[on background layer]
  \node[stage, fit={(0,0.55) (4.30,-3.05)}]     {};
  \node[stage, fit={(4.45,0.55) (9.60,-3.05)}]  {};
  \node[stage, fit={(9.75,0.55) (16.40,-3.05)}] {};
  \node[stage, fit={(0,-3.85) (11.05,-7.15)}]   {};
  \node[stage, fit={(11.20,-3.85) (16.40,-7.15)}] {};
\end{scope}
\node[tag] at (0.05,0.51)  {(a) Hypothesise \& simulate};
\node[tag] at (4.50,0.51)  {(b) Summarise \& compare};
\node[tag] at (9.80,0.51)  {(c) SMC-ABC refinement};
\node[tag] at (0.05,-3.89) {(d) Posterior representations};
\node[tag] at (11.25,-3.89){(e) Signature capability};

\node[font=\scriptsize] at (0.85,-0.28) {$y \sim \pi(y)$};
\node[hyp] (y1) at (0.85,-0.90) {OSCC};
\node[hyp] (y2) at (0.85,-1.47) {Mucocele};
\node[hyp] (y3) at (0.85,-2.04) {Atr.\ glossitis};

\node[llm, minimum height=11mm, minimum width=7mm] (llm) at (2.30,-1.47)
  {\rotatebox{90}{\textbf{LLM}}};
\node[font=\scriptsize\itshape, text=abcorange!75!black] at (1.95,-2.62)
  {class-conditional simulator};
\foreach \i in {1,2,3} \draw[ar] (y\i.east) -- (llm.west);

\node[card, text width=14mm] (sim) at (3.55,-1.47)
  {\textit{``firm ulcer, rolled margins, 6 weeks''}};
\draw[ar] (llm.east) -- (sim.west);
\node[font=\scriptsize] at (3.50,-0.42) {$x_{\mathrm{sim}}\!\sim\! p_{\mathrm{LLM}}(\cdot|y)$};

\node[font=\scriptsize] at (7.05,-0.20) {$\rho=1-\cos(z_{\mathrm{sim}},z_{\mathrm{obs}})$};
\node[card, minimum width=6mm, minimum height=4.4mm] (phi1) at (5.00,-0.95) {$\phi$};
\node[card, text width=15mm, draw=black!50, fill=black!4] (obs) at (5.25,-2.35)
  {\textbf{Observed}\\\textit{``painless indurated lesion, tongue''}};
\node[card, minimum width=6mm, minimum height=4.4mm] (phi2) at (6.55,-2.35) {$\phi$};
\draw[ar] (sim.east) to[out=0,in=180] (phi1.west);
\draw[ar] (obs.east) -- (phi2.west);

\begin{scope}[shift={(8.20,-1.35)}]
  \node[sub, minimum width=23mm, minimum height=21mm] {};
  \fill[black!80] (0.05,-0.38) circle (1.4pt);
  \node[font=\scriptsize, anchor=north west, inner sep=1pt] at (0.10,-0.42) {$z_{\mathrm{obs}}$};
  \foreach \x/\y in {-0.55/0.52, -0.24/0.40, -0.66/0.20, 0.42/0.55, 0.66/0.26}
    \fill[abcblue!45] (\x,\y) circle (1.2pt);
  \foreach \x/\y in {-0.24/-0.20, 0.30/-0.14, 0.02/0.06}
    \fill[abcblue!95] (\x,\y) circle (1.2pt);
  \draw[dashed, draw=abcgreen!85, line width=0.5pt] (0.05,-0.38) circle (0.62);
  \draw[<->, line width=0.4pt, draw=black!60] (0.05,-0.38) -- (-0.24,0.40);
  \node[font=\scriptsize, anchor=east, inner sep=1pt] at (-0.08,0.04) {$\rho$};
  \node[font=\scriptsize, anchor=north, text=abcgreen!50!black] at (0,-1.14) {accept if $\rho\le\epsilon$};
\end{scope}
\draw[ar] (phi1.east) to[out=0,in=180] (7.00,-0.95);
\draw[ar] (phi2.east) to[out=0,in=180] (7.00,-1.85);

\begin{scope}[shift={(11.05,-1.45)}]
  \foreach \dx/\r/\lab in {0/0.6/{$\epsilon_1$}, 1.85/0.38/{$\epsilon_2$}, 3.70/0.2/{$\epsilon_3$}} {
    \node[sub, minimum width=15.5mm, minimum height=18mm] at (\dx,-0.05) {};
    \draw[dashed, draw=abcgreen!85, line width=0.45pt] (\dx,0) circle (\r);
    \node[font=\scriptsize, text=abcgreen!55!black] at (\dx,-0.72) {\lab};
    \fill[black!80] (\dx,0) circle (1.1pt);
  }
  \foreach \x/\y in {-0.48/0.4,0.42/0.34,-0.4/-0.28,0.44/-0.36,0.1/0.12,-0.18/-0.48,0.55/0.02,-0.06/0.52,0.24/-0.14}
    \fill[abcblue!60] (\x,\y) circle (1.0pt);
  \foreach \x/\y in {1.60/0.24,1.90/0.18,1.66/-0.19,2.06/-0.21,1.98/0.04,1.78/0.31,2.11/0.09,1.82/-0.05}
    \fill[abcblue!78] (\x,\y) circle (1.0pt);
  \foreach \x/\y in {3.62/0.08,3.77/0.06,3.65/-0.08,3.79/-0.10,3.70/0.01,3.61/-0.02,3.75/0.13,3.67/0.12}
    \fill[abcblue!95] (\x,\y) circle (1.0pt);
  \draw[ar] (0.83,0) -- (1.02,0);
  \draw[ar] (2.68,0) -- (2.87,0);
  \node[font=\scriptsize, anchor=south] at (1.85,0.86)
    {reweight $\to$ resample $\to$ perturb};
  \node[font=\scriptsize, anchor=north, text=abcgreen!55!black] at (1.85,-0.95)
    {tightening tolerance $\epsilon_1\!>\!\epsilon_2\!>\!\epsilon_3$};
\end{scope}
\draw[ar] (9.35,-1.35) -- (10.05,-1.35);

\node[sub, minimum width=33mm, minimum height=25mm] at (1.90,-5.55) {};
\node[font=\scriptsize, text=abcblue!60!black] at (1.90,-4.55) {\msmc{}: \textbf{categorical}};
\begin{scope}[shift={(1.05,-6.45)}]
  \draw[line width=0.4pt, draw=black!50] (-0.22,0) -- (1.95,0);
  \foreach \i/\h/\l in {0/1.15/OC, 1/0.22/MC, 2/0.12/AG} {
    \fill[abcblue!65] (\i*0.70,0) rectangle (\i*0.70+0.42,\h);
    \node[font=\scriptsize, anchor=north, inner sep=1.5pt] at (\i*0.70+0.21,-0.02) {\l};
  }
\end{scope}
\node[font=\scriptsize] at (1.90,-4.92) {$\hat p(y\,|\,x_{\mathrm{obs}})$};

\node[sub, minimum width=33mm, minimum height=25mm] at (5.55,-5.55) {};
\node[font=\scriptsize, text=abcblue!60!black] at (5.55,-4.55) {\dsmc{}: \textbf{simplex}};
\begin{scope}[shift={(5.55,-6.45)}]
  \draw[line width=0.5pt, draw=black!55] (-1.0,0) -- (1.0,0) -- (0,1.12) -- cycle;
  \node[font=\scriptsize, anchor=north east, inner sep=0.5pt] at (-0.94,0.02) {OC};
  \node[font=\scriptsize, anchor=north west, inner sep=0.5pt] at (0.94,0.02) {MC};
  \node[font=\scriptsize, anchor=south west, inner sep=0.5pt] at (0.04,1.06) {AG};
  \foreach \x/\y in {-0.36/0.52,-0.27/0.62,-0.45/0.43,-0.19/0.47,-0.33/0.36,-0.48/0.56,-0.24/0.57,-0.38/0.62,-0.29/0.45}
    \fill[abcblue!75] (\x,\y) circle (1.1pt);
  \draw[dashed, draw=abcblue!55, line width=0.4pt] (-0.33,0.51) circle (0.26);
\end{scope}
\node[font=\scriptsize] at (5.55,-4.92) {$\hat p(\theta\,|\,x_{\mathrm{obs}})$};

\node[sub, minimum width=33mm, minimum height=25mm] at (9.20,-5.55) {};
\node[font=\scriptsize, text=abcblue!60!black] at (9.20,-4.55) {\abi{}: \textbf{amortised}};
\begin{scope}[shift={(8.30,-5.55)}]
  \node[font=\scriptsize, anchor=east, inner sep=1pt] at (0.02,0) {$z_{\mathrm{obs}}$};
  \foreach \dx/\n in {0.38/3, 0.82/4, 1.26/4} {
    \foreach \i in {1,...,\n} {
      \pgfmathsetmacro{\yy}{(\i-(\n+1)/2)*0.22}
      \fill[abcorange!75] (\dx,\yy) circle (1.1pt);
    }
  }
  \foreach \a/\na/\b/\nb in {0.38/3/0.82/4, 0.82/4/1.26/4} {
    \foreach \i in {1,...,\na} { \foreach \j in {1,...,\nb} {
      \pgfmathsetmacro{\ya}{(\i-(\na+1)/2)*0.22}
      \pgfmathsetmacro{\yb}{(\j-(\nb+1)/2)*0.22}
      \draw[line width=0.15pt, draw=abcorange!35] (\a,\ya) -- (\b,\yb);
    }}
  }
  \draw[ar] (0.10,0) -- (0.28,0);
  \draw[ar] (1.36,0) -- (1.56,0);
  \node[font=\scriptsize, anchor=west, inner sep=1pt] at (1.58,0) {$\mathrm{Dir}(\bm{\alpha})$};
  \node[font=\scriptsize, anchor=north, text=black!60] at (0.95,-0.58) {single forward pass,};
  \node[font=\scriptsize, anchor=north, text=black!60] at (0.95,-0.86) {no LLM calls};
\end{scope}
\node[font=\scriptsize] at (9.20,-4.92) {$\bm{\alpha}_\varphi(z_{\mathrm{obs}})$};

\draw[draw=black!45, line width=0.5pt]
  (13.00,-3.10) -- (13.00,-3.42) -- (5.55,-3.42) -- (5.55,-3.60);
\draw[draw=black!45, line width=0.5pt] (1.90,-3.60) -- (9.20,-3.60);
\foreach \x in {1.90, 5.55, 9.20} \draw[ar, draw=black!45] (\x,-3.60) -- (\x,-4.28);
\node[font=\scriptsize\itshape, text=black!55, anchor=south] at (3.72,-3.58) {or};
\node[font=\scriptsize\itshape, text=black!55, anchor=south] at (7.38,-3.58) {or};

\foreach \y in {-4.78, -5.70, -6.62}
  \draw[draw=black!25, fill=white, rounded corners=2pt]
    (11.40,\y-0.40) rectangle (16.28,\y+0.40);

\begin{scope}[shift={(11.92,-4.78)}]
  \draw[line width=0.35pt, draw=black!45] (-0.42,-0.30) -- (0.46,-0.30);
  \fill[abcblue!35] (-0.36,-0.30) rectangle (-0.14,-0.21);   %
  \fill[abcblue!85] (0.10,-0.30) rectangle (0.32,0.29);      %
  \draw[ar, draw=black!55] (-0.06,-0.10) -- (0.04,-0.10);
\end{scope}
\node[font=\scriptsize, anchor=west, align=left] at (12.58,-4.78)
  {\msmc{}: \textbf{prior injection}\\[1pt]
   \color{black!60}wrong prior: $0.14\!\to\!0.94$};

\begin{scope}[shift={(11.92,-5.70)}]
  \draw[line width=0.4pt, draw=black!50] (-0.40,-0.26) -- (0.40,-0.26) -- (0,0.30) -- cycle;
  \foreach \x/\y in {-0.14/-0.09,-0.04/-0.02,-0.18/-0.17,-0.02/-0.14,-0.10/-0.19,0.04/-0.09}
    \fill[abcblue!80] (\x,\y) circle (0.9pt);
  \draw[dashed, draw=abcblue!55, line width=0.35pt] (-0.07,-0.11) circle (0.17);
\end{scope}
\node[font=\scriptsize, anchor=west, align=left] at (12.58,-5.70)
  {\dsmc{}: \textbf{second-order}\\[1pt]
   \color{black!60}aleatoric $+$ epistemic split};

\begin{scope}[shift={(11.92,-6.62)}]
  \draw[line width=0.5pt, draw=abcorange!80] (0,0) circle (0.27);
  \draw[line width=0.5pt, draw=abcorange!80] (0,0) -- (0,0.17);
  \draw[line width=0.5pt, draw=abcorange!80] (0,0) -- (0.13,-0.06);
  \fill[abcorange!80] (0,0) circle (0.7pt);
\end{scope}
\node[font=\scriptsize, anchor=west, align=left] at (12.58,-6.62)
  {\abi{}: \textbf{throughput}\\[1pt]
   \color{black!60}$18\,\mu$s per case};

\foreach \y in {-4.78, -5.70, -6.62}
  \draw[ar, draw=black!40] (11.08,\y) -- (11.36,\y);

\end{tikzpicture}
\vspace{-2pt}
\caption{\textbf{Closed-set clinical classification as likelihood-free
posterior inference.} \textbf{(a)}~A diagnosis $y$ is drawn from the
prior $\pi(y)$ and the prompt-conditioned LLM is run \emph{as a
class-conditional simulator}, generating $x_{\mathrm{sim}}$ for that
hypothesis. \textbf{(b)}~Simulated and observed descriptions are mapped
to a shared embedding space by $\phi$ and compared by cosine distance
$\rho$; a particle is accepted when $\rho\le\epsilon$. Everything
compared against $z_{\mathrm{obs}}$ is simulated: in \msmc{} it is a
fresh simulation from the particle's own class, in \dsmc{} it is
$\tilde z(\theta)$, a $\theta$-weighted combination of the $K$ class-mean
simulated embeddings. \textbf{(c)}~SMC-ABC repeats
reweight--resample--perturb under a tightening tolerance
$\epsilon_1>\epsilon_2>\epsilon_3$. Note the arrow direction: the LLM is
never asked to score $x_{\mathrm{obs}}$, only to generate under a
hypothesis, which is what makes the intractable likelihood unnecessary.
\textbf{(d)}~Three alternative representations of the posterior, not
stages of a pipeline: \msmc{} places particles on class labels and uses
no class means; \dsmc{} places them on the simplex and scores them
through $\tilde z(\theta)$; \abi{} amortises the \dsmc{} posterior into a
single forward pass. \textbf{(e)}~The capability that
\emph{distinguishes} each, which is what decides between them; the full
matrix is Table~\ref{tab:variants}. Both \msmc{} and \dsmc{} accept a
clinician prior at test time, and \msmc{} is the more robust of the two
when that prior is wrong (\S\ref{sec:prior}). Class codes: OC = oral
squamous cell carcinoma, MC = mucocele, AG = atrophic glossitis.}
\label{fig:schematic}
\end{figure*}

\section{Datasets}
\label{sec:datasets}

We evaluate on three clinical benchmarks spanning synthetic
expert-validated content and real published case reports.
\paragraph{\olb{} (released with this work).}
\olb{} contains 300 vignettes over three phenotypically overlapping but
clinically distinct conditions, Oral Squamous Cell Carcinoma (OSCC),
Oral Mucocele (MC) and Atrophic Glossitis (AG), with $N{=}100$ per
class. Vignettes were synthesised from authoritative oral pathology
references \citep{neville2016oral,regezi2016oral,who2022headneck} to
preserve phenomenological accuracy, then reviewed blind by an oral
pathologist who scored each case on two five-point Likert dimensions:
\emph{Completeness} (are the necessary diagnostic features present:
morphology, temporal evolution, anatomic site) and \emph{Correctness}
(is the symptom constellation medically plausible). Mean scores were
$4.86/5$ ($\sigma{=}0.45$) and $4.55/5$ ($\sigma{=}0.80$), with reviewer
feedback used to refine borderline cases. Because Completeness asks the
reviewer to confirm that the features supporting the gold label are
present, \olb{} presents well-specified evidence and so isolates
calibration under clear evidence, while \multicare{} supplies the
higher-ambiguity setting. Per-class statistics, example vignettes and
the annotation protocol are in Appendix~\ref{app:datasets}.
\paragraph{\multicare{}.}
To evaluate beyond synthetic vignettes, we use
\multicare{}~\citep{nievas_offidani_2023_10079370}, a corpus of real
published clinical case reports. We retain seven clinically distinct
classes covering common acute and chronic presentations, with the full
class list and preprocessing details provided in
Appendix~\ref{app:datasets}.

\paragraph{Symptom-to-Diagnosis (\sd{}).}
We additionally evaluate on the public GretelAI Symptom-to-Diagnosis
dataset~\citep{gretelai2023symptomtodiagnosis}, restricted to six
common diagnostic classes: arthritis, bronchial asthma, jaundice,
drug reaction, cervical spondylitis, and dengue.

\section{Experimental Setup}
\label{sec:setup}

\paragraph{Models.}
We evaluate six open LLMs from the Llama, Mistral, and Qwen model
families, spanning instruction-tuned models and clinical fine-tuned in the 7--8B parameter range:
Mistral-7B-Instruct-v3~\citep{jiang2023mistral},
Llama-3.1-8B-Instruct~\citep{grattafiori2024llama3},
Llama3-Med42-8B~\citep{christophe2024med42},
Qwen-2.5-7B-Instruct~\citep{yang2024qwen2},
Microsoft-Orca-2-7B~\citep{mitra2023orca}, and
AlpaCare-LLaMA-7B~\citep{zhang2023alpacare}.
Decoding temperature is fixed at $T=0.2$ for the main results
(temperature-sensitivity ablation in Appendix~\ref{app:temperature}).

\paragraph{Embedding model.}
For semantic distance computation we use OpenAI
\texttt{text-embedding-3-large}. We additionally evaluate five
alternative embeddings (general-purpose and biomedical) in
Appendix~\ref{app:embeddings}.

\paragraph{Baselines.}
We compare against four uncertainty estimation baselines spanning the
heuristic and consistency families:
\textbf{Model Logits (ML)} extracts the softmax probability assigned to
the selected option in a constrained multiple-choice prompt.
\textbf{Elicited Probability (EP)}~\citep{xiong2024can} prompts the
model to verbalise its top-$k$ guesses with self-reported probabilities.
\textbf{Prompt Ensemble (PE)} aggregates predictions across $K{=}8$
semantically equivalent prompt templates (templates in
Appendix~\ref{app:prompts}).
\textbf{Elicited Fidelity (EF)}~\citep{zhang2024calibrating} estimates
uncertainty as the entropy of the empirical class distribution under
repeated stochastic decoding.

\paragraph{Metrics.}
We report \textbf{Accuracy} and \textbf{Macro-F1} for predictive
performance, and \textbf{Brier
score}, \textbf{Expected Calibration Error} (ECE; 10 bins), and
\textbf{predictive Entropy} for calibration and uncertainty quality.
For selective-prediction analysis (\S\ref{sec:selective}) we additionally
report risk--coverage curves~\citep{geifman2017selective}.

\section{Main Results}
\label{sec:results}
\S\ref{sec:first-order} and \S\ref{sec:distribution-aware} compare
\msmc{} against scalar-confidence baselines and the distribution-aware
methods respectively. \S\ref{sec:error-detection} then asks whether the
inferred uncertainty identifies errors, and \S\ref{sec:embedding-ablation}
how much of the benefit survives once the embedding baseline is itself
calibrated; in our view these two decide whether the framework is worth
its cost. \S\ref{sec:selective}--\ref{sec:ood} evaluate selective
prediction, prior sensitivity, and out-of-distribution behaviour.
Second-order uncertainty, SMC dynamics, and significance testing are in
Appendices~\ref{sec:second-order}, \ref{app:smc-dynamics}
and~\ref{sec:ablation-significance}.

\subsection{First-Order Uncertainty Quantification}
\label{sec:first-order}
We first evaluate \msmc{} against three scalar-uncertainty baselines:
Model Logits (ML), Elicited Probability (EP) and Prompt Ensemble (PE).
Table~\ref{tab:main} reports accuracy, F1, ECE, Brier and entropy across
six base models and three benchmarks. \msmc{} improves both accuracy and
calibration across the grid, and the margin is largest where the base
model is weakest: averaged over the six LLMs on \multicare{} it reaches
$96.0\%$ accuracy against $46.4$--$52.0\%$ for the baselines, a gap of
roughly $44$ points. The individual cells behind that average are striking
(per-model results in Appendix~\ref{app:full-tables},
Table~\ref{tab:main-full}): with AlpaCare-7B, Model Logits collapse to
$5.4\%$ accuracy at $0.91$ ECE, near-random predictions held with
near-total confidence, while \msmc{} attains $97.0\%$ accuracy at $0.07$
ECE on the identical inputs, a $17\times$ improvement in accuracy and a
$13\times$ reduction in calibration error. Orca follows the same pattern
(Table~\ref{tab:main-full}). On strong base models
the margin narrows, and on \olb{} with Llama-3.1 \msmc{} trades
accuracy ($90.6$ vs $98.6$) for better calibration ($0.04$ vs $0.12$
ECE). The pattern is consistent: the method contributes most
exactly where the base model's own confidence is least trustworthy.

\subsection{Beyond First Order: Distribution-Aware Methods}
\label{sec:distribution-aware}
The baselines above return a single confidence scalar. We now compare
methods that return a distribution: Elicited Fidelity (EF), which
aggregates repeated stochastic generations into an empirical class
distribution and its entropy, against \dsmc{} and \abi{}, which return
parametric Bayesian posteriors over the simplex
(Table~\ref{tab:advanced-uq}).

On classification, \dsmc{} and \abi{} match or exceed EF across most
model--dataset combinations, with the gap widening on harder benchmarks.
\dsmc{} gives the strongest Brier scores; ECE is more mixed, with EF
sharp on \olb{}, where consistency closely tracks accuracy, but weaker on
\sd{} and \multicare{}.

The methods differ sharply in cost. EF and \dsmc{} both require repeated
LLM generations per prediction, at $3.7$--$6.9$\,s and $15$--$32$\,s per
patient respectively. \abi{} replaces per-instance generation with an
embedding lookup and a single forward pass at $0.13$--$0.15$\,s
end-to-end; where embeddings are precomputed the forward pass alone takes
$16$--$19\,\mu$s, \textbf{five orders of magnitude below EF and at higher
mean accuracy on two of three benchmarks}. Amortisation moves the
dominant cost to a one-off training run while preserving the Dirichlet
structure used for second-order uncertainty
(Appendix~\ref{sec:second-order}), which makes uncertainty-aware triage
tractable at clinical throughput.

\abi{} is dataset-dependent. On \olb{}, where class-conditional simulator
outputs cluster tightly, its Dirichlet posterior becomes
over-concentrated (median $\alpha_0>500$), a known difficulty for neural
posterior estimation under low-stochasticity simulators
\citep{pmlr-v202-bengs23a}; on \sd{} and \multicare{} it settles at
$\alpha_0\in[100,150]$ (Appendix~\ref{sec:abi-details}). Fidelity to its
\dsmc{} target is nonetheless high on individual cases: across all $669$
\multicare{} cases, top-class agreement is $97\%$ and posterior means
correlate at $r=0.975$ (Table~\ref{tab:abi-fidelity}). Concentration is
the exception, so \dsmc{} remains the variant to use when second-order
quantities are themselves the object of interest.

\smallskip
\noindent\emph{Choosing among the variants.} The three are
complementary operating points, not interchangeable implementations, and
no single one provides every capability we claim for the framework
(Table~\ref{tab:variants} in Appendix~\ref{app:full-tables} states this
capability by capability). Two exclusions matter most: \abi{} amortises
a \emph{fixed} training prior, so the prior-conditioned inference of
\S\ref{sec:prior} is unavailable in the deployment variant; and the
second-order decomposition requires a distribution over the simplex, so
it is unavailable in \msmc{}. \msmc{} is the choice when an
interpretable posterior over diagnoses and test-time priors matter,
\dsmc{} when second-order uncertainty does, \abi{} when throughput
does.

\begin{table*}[t]
\centering
\footnotesize
\setlength{\tabcolsep}{3.5pt}
\renewcommand{\arraystretch}{1.0}
\begin{tabular}{l cccc || cccc || cccc}
\toprule
& \multicolumn{4}{c}{\olb{}} & \multicolumn{4}{c}{\sd{}} & \multicolumn{4}{c}{\multicare{}} \\
\cmidrule(lr){2-5}\cmidrule(lr){6-9}\cmidrule(lr){10-13}
Method & Acc$\uparrow$ & F1$\uparrow$ & Brier$\downarrow$ & ECE$\downarrow$
       & Acc$\uparrow$ & F1$\uparrow$ & Brier$\downarrow$ & ECE$\downarrow$
       & Acc$\uparrow$ & F1$\uparrow$ & Brier$\downarrow$ & ECE$\downarrow$ \\
\midrule
ML & 84.1 & 83.2 & \best{0.072} & 0.108 & 51.9 & 47.7 & 0.565 & 0.190 & 46.4 & 43.8 & 0.848 & 0.333 \\
EP & 61.7 & 53.2 & 0.283 & 0.250 & 63.9 & 60.2 & 0.588 & \best{0.125} & 51.9 & 46.8 & 0.615 & 0.157 \\
PE & 87.2 & 86.7 & 0.137 & 0.085 & 61.8 & 59.4 & 0.595 & 0.185 & 52.0 & 47.5 & 0.688 & 0.213 \\
\msmc{} (ours) & \best{95.5} & \best{95.4} & 0.030 & \best{0.023} & \best{81.7} & \best{79.7} & \best{0.228} & 0.110 & \best{96.0} & \best{96.1} & \best{0.068} & \best{0.032} \\
\bottomrule
\end{tabular}
\caption{Main results, \textbf{averaged over the six base LLMs}.
Baselines are \textbf{ML}, Model Logits (softmax probability of the
selected option); \textbf{EP}, Elicited Probability (the model's own
verbalised confidence); and \textbf{PE}, Prompt Ensemble (agreement
across $K{=}8$ paraphrased prompts); all three are defined in
\S\ref{sec:setup}. Bold marks the best value per metric within each
dataset. \msmc{} improves
mean accuracy and calibration on all three benchmarks, most
dramatically on \multicare{}, where the scalar-confidence baselines
average below $52\%$ accuracy. Averages hide substantial spread, which
is itself informative: \msmc{} accuracy ranges over $90.6$--$98.3$ on
\olb{} while ML ranges over $65.3$--$98.6$, so the method is not only
better on average but markedly more stable across base models.
Per-model results, including the cells where a baseline wins, are
given in full in Appendix~\ref{app:full-tables},
Table~\ref{tab:main-full}.}
\label{tab:main}
\end{table*}

\begin{table*}[t]
\centering
\footnotesize
\setlength{\tabcolsep}{3.5pt}
\renewcommand{\arraystretch}{1.0}
\begin{tabular}{l cccc || cccc || cccc}
\toprule
& \multicolumn{4}{c}{\olb{}} & \multicolumn{4}{c}{\sd{}} & \multicolumn{4}{c}{\multicare{}} \\
\cmidrule(lr){2-5}\cmidrule(lr){6-9}\cmidrule(lr){10-13}
Method & Acc$\uparrow$ & Brier$\downarrow$ & ECE$\downarrow$ & Lat.$\downarrow$
       & Acc$\uparrow$ & Brier$\downarrow$ & ECE$\downarrow$ & Lat.$\downarrow$
       & Acc$\uparrow$ & Brier$\downarrow$ & ECE$\downarrow$ & Lat.$\downarrow$ \\
\midrule
EF             & 90.3 & 0.162 & \best{0.070} & 6\,s
               & 68.1 & 0.573 & 0.260 & 8\,s
               & 76.7 & 0.413 & \best{0.187} & 4\,s \\
\dsmc{} (ours) & 90.8 & \best{0.080} & 0.262 & 15\,s
               & \best{81.2} & \best{0.390} & \best{0.222} & 18\,s
               & 95.8 & \best{0.127} & 0.195 & 25\,s \\
\abi{} (ours)  & \best{93.7} & 0.222 & 0.278 & \best{0.13\,s}
               & 81.0 & 0.432 & 0.300 & \best{0.15\,s}
               & \best{97.5} & 0.155 & 0.278 & \best{18\,$\mu$s} \\
\bottomrule
\end{tabular}
\caption{Distribution-aware methods, \textbf{averaged over the six base
LLMs}. \textbf{EF} is Elicited Fidelity, the consistency baseline that
aggregates repeated stochastic generations into an empirical class
distribution (\S\ref{sec:setup}). Latency is per patient, rounded. \dsmc{} gives the best mean
Brier on all three benchmarks and \abi{} the best mean accuracy on two,
EF retains an ECE advantage on \olb{} and \multicare{}, where
consistency tracks accuracy closely; its errors are nonetheless poorly
ranked (AUROC $0.584$, \S\ref{sec:error-detection}), which is the
property abstention depends on.
\abi{} converts a $15$--$25$\,s per-patient cost into $0.13$\,s (or
$18\,\mu$s where embeddings are precomputed) after a one-off training
run. Per-model results in Appendix~\ref{app:full-tables},
Table~\ref{tab:advanced-uq-full}.}
\label{tab:advanced-uq}
\end{table*}

\subsection{Does the Uncertainty Identify Errors?}
\label{sec:error-detection}

Calibration metrics such as ECE summarise agreement between confidence
and accuracy in aggregate; they do not establish that the uncertainty
is useful for deciding \emph{which individual predictions} to trust.
That property is what a deployed system actually needs, so we measure it
directly: we treat misclassification as the positive class, rank
predictions by predictive entropy, and report detection AUROC across all
six base models on \multicare{}.

Table~\ref{tab:error-detection} gives the result. \msmc{} exceeds
$0.91$ AUROC on every base model (mean $0.923$), against $0.760$ for
Model Logits and $0.584$ for EF. The EF figure is worth dwelling on: EF
is the \emph{most accurate} of the four baselines ($0.767$), yet its
entropy is close to uninformative about which of its own predictions are
wrong. That is precisely the failure this paper set out to address -
a method can be accurate and consistent while its confidence signal
carries almost nothing.

Because AUROC can flatter a method that is simply more accurate, we also
report excess-AURC, which subtracts the AURC of a discrete-optimal
ranking at the same error count and therefore measures ranking quality
alone. \textbf{The margin widens rather than closes}: \msmc{} and
\abi{} reach E-AURC of $0.0035$--$0.0037$ against $0.11$--$0.17$ for
every baseline, a separation of roughly \textbf{thirty-fold} on a
metric from which accuracy has been removed by construction. All $24$
\msmc{}-versus-baseline comparisons favour \msmc{} on both AUROC and
E-AURC, with patient-level bootstrap intervals excluding zero.

\begin{table}[t]
\centering
\footnotesize
\setlength{\tabcolsep}{4pt}
\renewcommand{\arraystretch}{1.05}
\begin{adjustbox}{max width=\columnwidth}
\begin{tabular}{l ccccc}
\toprule
Method & Acc. & AUROC$\uparrow$ & AUROC range & E-AURC$\downarrow$ & Risk@80\%$\downarrow$ \\
\midrule
\msmc{} (ours) & 0.960 & \best{0.923} & 0.912--0.959 & 0.0037 & \best{0.0062} \\
\abi{} (ours)  & \best{0.975} & 0.883 & 0.864--0.910 & \best{0.0035} & 0.0096 \\
\dsmc{} (ours) & 0.958 & 0.875 & 0.829--0.923 & 0.0075 & 0.0115 \\
\midrule
ML & 0.464 & 0.760 & 0.686--0.805 & 0.1096 & 0.4794 \\
EP & 0.536 & 0.695 & 0.554--0.805 & 0.1515 & 0.4009 \\
PE & 0.520 & 0.694 & 0.564--0.756 & 0.1673 & 0.4246 \\
EF & 0.767 & 0.584 & 0.495--0.644 & 0.1534 & 0.2106 \\
\bottomrule
\end{tabular}
\end{adjustbox}
\caption{\textbf{Does the uncertainty identify errors?} Misclassification
detection on \multicare{}, averaged over all six base LLMs, using
predictive entropy as the ranking score and misclassification as the
positive class. AUROC range gives the min--max across the six models.
E-AURC is excess area under the risk--coverage curve (AURC minus the
AURC of a discrete-optimal ranking at the same error count), which
removes the contribution of full-coverage accuracy and therefore
isolates \emph{ranking} quality rather than rewarding an accurate
classifier. Risk@80\% is selective risk at 80\% coverage. All $24$
\msmc{}-versus-baseline comparisons favour \msmc{} on both AUROC and
E-AURC, with patient-level bootstrap confidence intervals excluding
zero. The E-AURC column carries the main point: because it removes the
contribution of full-coverage accuracy, the $\sim\!30\times$ margin it
shows is a difference in \emph{ranking} quality, not a by-product of
our methods being more accurate. Reported over the six-model
\multicare{} grid.}
\label{tab:error-detection}
\end{table}

\subsection{Is the Benefit From the Bayesian Layer or the Embedding?}
\label{sec:embedding-ablation}

The framework compares texts in an embedding space, so it is worth
isolating how much of its behaviour the representation alone accounts
for. Raw normalised cosine scores are poorly calibrated (Brier $0.80$,
ECE $0.78$; Appendix~\ref{app:cosine}), but the more demanding
comparison gives that baseline the post-hoc calibration a practitioner
would apply. We fit a centroid-cosine classifier over the same
embeddings with scalar temperature scaling (TS) on a group-stratified
$20\%$ calibration split, evaluating once on the untouched $80\%$ of
Mistral/\multicare{} and using the identical split for every method
(Table~\ref{tab:calibrated-embedding}). Given a labelled calibration
set, that classifier is itself well calibrated, reaching ECE $0.0145$
against $0.0166$ for native \msmc{}: calibration here is attainable by
either route.

\begin{table}[t]
\centering
\footnotesize
\setlength{\tabcolsep}{4.5pt}
\renewcommand{\arraystretch}{1.05}
\begin{adjustbox}{max width=\columnwidth}
\begin{tabular}{l ccccc}
\toprule
Method & Acc.$\uparrow$ & NLL$\downarrow$ & Brier$\downarrow$ & ECE$\downarrow$ & AUROC$\uparrow$ \\
\midrule
Centroid cosine + TS   & 0.964 & 0.108 & 0.054 & \best{0.0145} & \best{0.950} \\
\midrule
\msmc{} (native)       & 0.957 & 0.168 & 0.068 & 0.0166 & 0.922 \\
\msmc{} + TS           & 0.957 & 0.167 & 0.067 & 0.0254 & 0.924 \\
\dsmc{} + TS           & 0.966 & 0.157 & 0.062 & 0.0350 & 0.902 \\
\abi{} + TS            & \best{0.983} & 0.112 & \best{0.031} & 0.0165 & 0.934 \\
\bottomrule
\end{tabular}
\end{adjustbox}
\caption{\textbf{Does the Bayesian layer add anything over a calibrated
embedding classifier?} Mistral-7B on \multicare{}. A centroid-cosine
classifier over the same embeddings, with scalar temperature scaling
(TS) fitted on a group-stratified 20\% calibration split and evaluated
once on the untouched 80\%; the identical split is used for every
method. Given the same post-hoc calibration and a labelled calibration
set, the embedding classifier is itself well calibrated
(ECE $0.0145$), comparable to native \msmc{} at $0.0166$ --- so
calibration in this regime is attainable by either route. Under matched
supervision the framework goes further: \textbf{\abi{}+TS attains the
best accuracy ($0.983$ vs $0.964$) and the best Brier score ($0.031$ vs
$0.054$) in the table}. \msmc{} additionally reaches its calibration
\emph{natively}, without any labelled calibration split, and the
posterior supports prior injection and second-order uncertainty, which
a similarity classifier does not provide at any level of supervision.}
\label{tab:calibrated-embedding}
\end{table}

Three things follow. First, under matched supervision the framework
\emph{exceeds} the calibrated embedding baseline outright: \abi{}+TS
attains higher accuracy ($0.983$ vs $0.964$) and lower Brier ($0.031$ vs
$0.054$), so the posterior is doing more than recovering embedding-level
performance. Second, \msmc{} reaches comparable calibration
\emph{natively}, with no labelled calibration split, the operative
constraint whenever no held-out labelled data exist for the target
label set, as for a new or rare diagnostic panel. Third, and
unavailable to any similarity classifier at any level of supervision,
the posterior accepts a test-time clinical prior (\S\ref{sec:prior})
and, in the Dirichlet variants, decomposes uncertainty into aleatoric
and epistemic parts (Appendix~\ref{sec:second-order}). Temperature
scaling is complementary to the framework rather than an alternative:
\dsmc{} and \abi{} both improve under it. We recommend the calibrated
embedding classifier as a standard baseline for this setting, since a
comparison against uncalibrated similarity scores overstates what the
inference layer contributes.

\subsection{Selective Prediction Analysis}
\label{sec:selective}

\S\ref{sec:error-detection} established error ranking in aggregate. We
now show what it means operationally, via risk--coverage
curves~\citep{geifman2017selective}: predictions are ranked by
predictive entropy, and we report risk ($1-\mathrm{accuracy}$) on the
top-$c$ fraction of most-confident predictions.

\begin{figure*}[t]
\centering
\begin{subfigure}[b]{0.442\linewidth}
\includegraphics[width=\linewidth]{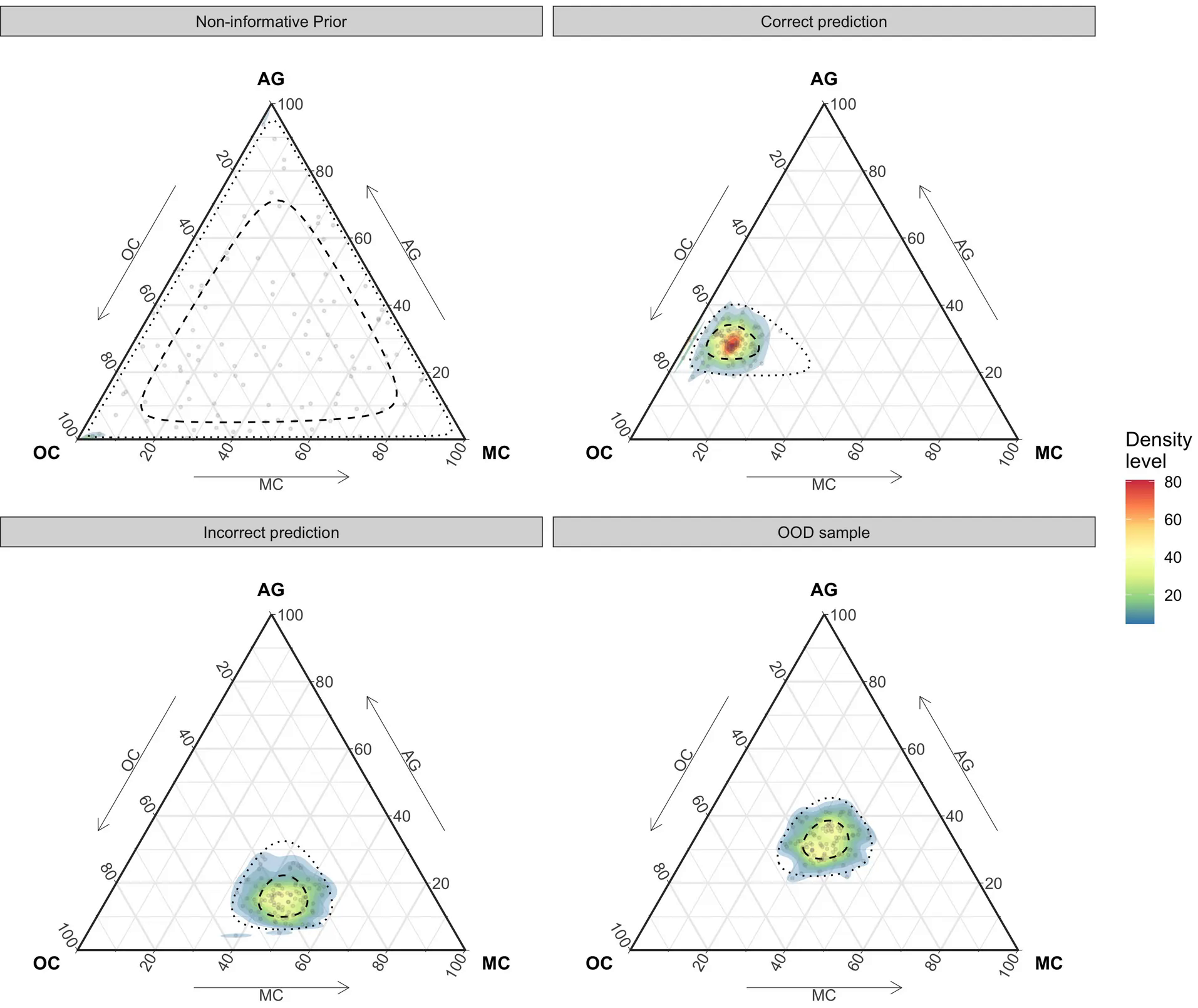}
\caption{\dsmc{} posteriors on the \olb{} simplex}
\label{fig:ternary}
\end{subfigure}\hfill
\begin{subfigure}[b]{0.528\linewidth}
\includegraphics[width=\linewidth]{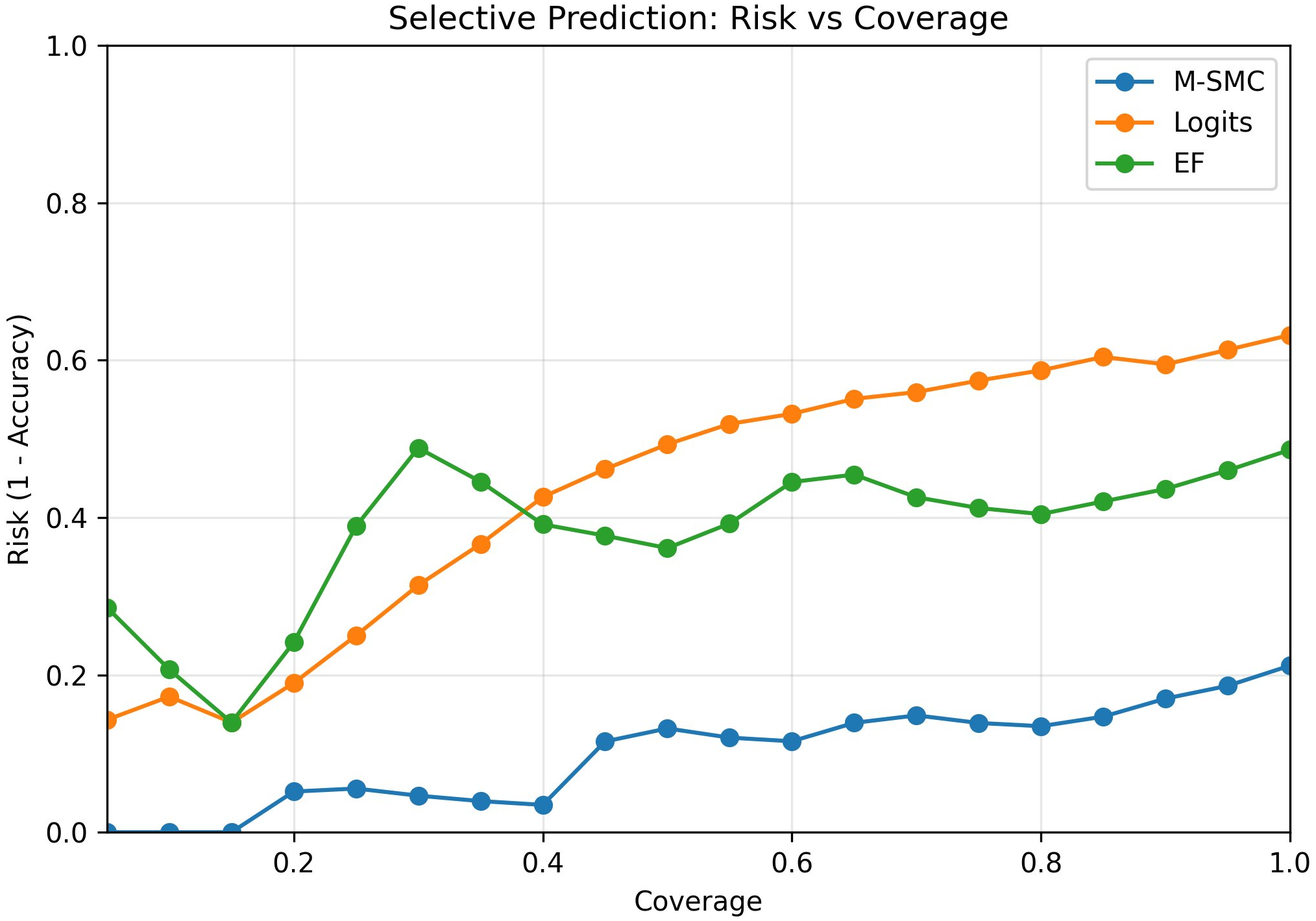}
\caption{Selective risk on \sd{}}
\label{fig:selective}
\end{subfigure}
\caption{\textbf{Posterior geometry and its operational consequence.}
\textbf{(a)}~\dsmc{} posteriors on the \olb{} probability simplex
(Mistral-7B); vertices are OSCC (OC), mucocele (MC) and atrophic
glossitis (AG). Clockwise from top left: a non-informative prior is
diffuse; a correctly classified case concentrates near the true vertex
(here AG); an OOD input stays diffuse near the simplex centre,
expressing the maximal uncertainty appropriate to an input outside the
label space; and a misclassified case concentrates on the wrong region,
the confident-error mode analysed in
Appendix~\ref{app:error-structure}.
\textbf{(b)}~Risk--coverage behaviour on \sd{} with AlpaCare-7B.
Predictions are ranked by predictive entropy and risk
($1-\mathrm{accuracy}$) is reported on the most-confident fraction. At
$80\%$ coverage \msmc{} attains risk $0.13$ against $0.40$ for EF and
$0.59$ for logits. We plot risk only, since accuracy--coverage is its
exact complement and adds no information. Read together, the two panels
make the same point at different granularities: posterior mass
concentrates only when the evidence warrants it, which is what makes
entropy-based abstention work.}
\label{fig:posterior-geometry}
\end{figure*}

Figure~\ref{fig:selective} shows the result for \sd{} with
AlpaCare-7B (the model on which heuristic baselines fail most
severely). \msmc{} dominates both EF and logits at every coverage
level: at $80\%$ coverage, \msmc{} risk is $0.13$ versus $0.40$ (EF)
and $0.59$ (logits).

The vertical gap partly reflects a difference in full-coverage accuracy
($0.79$, $0.51$, $0.37$), so it is not by itself evidence about
uncertainty quality; the \emph{shape} of the curves is. A score that
ranks well produces risk increasing monotonically as coverage grows.
\msmc{} does, rising smoothly from $0.00$ at low coverage to $0.21$ at
full coverage. EF does not: its risk oscillates between $0.14$ and
$0.49$ over the first half of the coverage range, so abstaining under EF
discards roughly as many correct predictions as errors. Stratifying by entropy shows the same effect at
matched coverage: \msmc{} places $88.5\%$ of cases in its
high-confidence bin at $83.9\%$ accuracy, whereas EF places $77.4\%$
there at only $59.6\%$, and logits place just $5.9\%$ there, leaving
$70\%$ of cases in the high-risk low-confidence bin
(Appendix~\ref{app:full-tables}, Table~\ref{tab:stratified}).

\subsection{Prior Sensitivity Analysis}
\label{sec:prior}

If the method performs Bayesian updating, then an informative
prior on the correct class should require only a small correction, a
misleading prior should be recoverable as evidence accumulates, and
neither behaviour should be available to similarity-based methods, which
have no prior to update. We test all three.

We run \msmc{} on \olb{} under a uniform $\mathrm{Dir}(1,1,1)$ prior, an
informative $\mathrm{Dir}(5,1,1)$ ($71.4\%$ mass on the true class), and
a misleading $\mathrm{Dir}(1,1,5)$ ($71.4\%$ on an incorrect class),
measuring true-class posterior mass before and after inference.
Table~\ref{tab:prior} establishes all three claims for \msmc{}. Informative priors lead to small posterior corrections
(true-class $\Delta = +0.26$), reflecting that the posterior largely
confirms a prior that already pointed the right way. Crucially, under a
\emph{misleading} prior, the posterior corrects the wrong belief: the
true-class probability shifts from $0.14$ to $0.94$
($\Delta = +0.79$) and accuracy is preserved at $97\%$. This
recovery is the signature of genuine evidence accumulation. It is not
merely stronger than the baselines: no score-based, elicitation-based or
consistency-based method can exhibit this behaviour \emph{at all},
because none of them represents a prior that could be wrong.

\dsmc{} updates in the right direction under every prior but by smaller
margins ($\Delta = +0.24$, $+0.09$, $+0.16$), and under the misleading
prior its accuracy falls to $0.19$. The reason is structural: its
particles are compared to $z_{\mathrm{obs}}$ through the mixture
$\tilde z(\theta)=\sum_k \theta_k \bar z_k$, so a prior placing mass away
from the true vertex biases which region of the simplex is explored, and
the mixture discrepancy is too smooth to pull particles back within the
iteration budget. \msmc{}, which resamples discrete hypotheses and
re-simulates under each, has no such barrier. We therefore recommend
\msmc{} whenever a clinician-supplied prior may be wrong.

These results are for \olb{}, a three-class setting in which the
likelihood is strong; how far prior recovery extends to label sets with
weaker per-class separation is an open question we take up in
\S\ref{sec:limitations}.

\subsection{Out-of-Distribution Behaviour}
\label{sec:ood}

We constructed expert-validated OOD variants of \olb{} and \sd{}
consisting of clinically plausible but out-of-class descriptions
(for \olb{}: dental complaints outside the three lesion classes,
e.g.\ denture instability, TMJ pain, gingival bleeding; for \sd{}:
unrelated symptom narratives such as migraine and gastrointestinal
complaints). We do \emph{not} use semantic-nonsense distractors,
which are trivial to flag; the test is whether the posterior remains
appropriately diffuse when the input does not correspond to any
modelled class.

Figure~\ref{fig:ternary} shows \dsmc{} posteriors on the \olb{} simplex
for four representative cases: a non-informative prior, a correct
classification, a misclassification, and an OOD input.

Quantitatively, Table~\ref{tab:ood} reports predictive entropy on the
OOD variants. On \sd{}-OOD, \msmc{} attains essentially the
\emph{maximum attainable} predictive entropy ($H\approx 2.5$ of
$\log_2 6 = 2.58$ bits) on all four models, the ideal response to an
input belonging to no modelled class, whereas logits retain confidence
($1.0$--$2.1$~bits) and EP is lower still ($0.7$--$1.2$~bits). \dsmc{}
is more conservative at $1.5$--$1.8$~bits, consistent with its smoother
simplex representation. Cosine distance in a general-purpose embedding space need not capture
every clinically salient distinction, so maximal entropy on every OOD
case is not guaranteed; the observed shift towards diffuse posteriors is
nonetheless in the right direction and well beyond the baselines.

\section{Discussion and Conclusion}
\label{sec:discussion}
\paragraph{Why posterior inference matters.}
The framework is best understood as a posterior-based decision layer,
not a post-hoc confidence estimator for an unchanged LLM output. That
distinction explains why accuracy can change at all: the decision is
the MAP of the inferred posterior. It also shifts the relevant
question from ``is the confidence calibrated?'' to ``does the posterior
rank errors correctly?''. On that question the evidence is clear
(\S\ref{sec:error-detection}): AUROC $0.923$ against $0.760$ for
softmax and $0.584$ for consistency, holding under excess-AURC.

\paragraph{Beyond semantic matching.}
The method uses embedding similarity as the ABC discrepancy, so it is
fair to ask what the inference layer adds over a nearest-prototype
classifier. Calibration alone does not separate them: given a labelled
calibration split, a temperature-scaled centroid-cosine classifier is
also well calibrated (\S\ref{sec:embedding-ablation}). Three
properties do. The posterior aggregates evidence across stochastic
class-conditional simulations rather than committing to the nearest
centroid, and this is not vacuous: true-class mass rises from $0.24$ to
$0.76$ across populations (Appendix~\ref{app:smc-dynamics}), so the
answer at convergence is not the answer at initialisation. It updates an
explicit prior, recovering the true class on \olb{} even from a
misleading one, which no similarity-only method can do because none
represents a prior at all. And it yields a distribution over the
simplex, supporting the aleatoric--epistemic split
(Appendix~\ref{sec:second-order}). The prior mechanism is the one we
would emphasise clinically: risk factors, prevalence, and patient
history routinely supply beliefs that should be combined with textual
evidence rather than discarded.

\paragraph{Conclusion.}
Uncertainty for closed-set clinical classification is better obtained by
inference than by attaching a score to a fixed prediction. Treating a
prompted LLM as a class-conditional simulator makes the intractable
likelihood unnecessary, and SMC-ABC turns the resulting simulations into
a posterior from which the prediction and its uncertainty are read off
together. Across three benchmarks and six open LLMs that posterior
separates correct predictions from errors far more reliably than
softmax, verbalised or consistency-based confidence, on a metric that
controls for accuracy, and it supports two things a confidence score
cannot: an explicit clinician prior, from which the true diagnosis is
recovered even when the prior points the wrong way, and a distribution
over the simplex separating ambiguity from unreliability. The three
variants span high-fidelity per-case inference to single-forward-pass
deployment at $18\,\mu$s. What the framework does not provide is
protection against a bias shared by its simulator and its embedding; we
regard detecting that bias, and composing the posterior with conformal
coverage guarantees, as the natural next steps. We release \olb{} to
support further work on uncertainty quantification in clinical NLP.

\section{Limitations}
\label{sec:limitations}

\paragraph{Computational cost.}
\msmc{} and \dsmc{} evaluate multiple class hypotheses per input, so
latency scales with the simulation budget ($15$--$25$\,s per patient
against milliseconds for a softmax score). \abi{} amortises this into a
single forward pass, and simulations can be cached aggressively: a
pool of three descriptions per class already reaches $95.6\%$ accuracy
(Appendix~\ref{app:pool}). SMC-ABC nonetheless remains suited to
offline analysis or high-stakes cases where latency is acceptable.

\paragraph{Simulator and embedding dependence.}
The posterior is defined under a prompt-conditioned simulator and an
embedding summary, not under the true clinical data-generating process,
so it inherits any infidelity in either. Embedding choice matters
substantially: general-purpose encoders outperform several biomedical
ones (Appendix~\ref{app:embeddings}) and should be treated as
summary-statistic selection in ABC, not an engineering detail.
Robustness cuts the other way for quantisation: under 4-bit AWQ,
logit-based confidence degrades (AUROC $98.1\to93.5$) while \msmc{}
reaches AUROC $100.0$ (Appendix~\ref{app:quantization}), because
inference runs on generated descriptions rather than token
probabilities.

\paragraph{Semantic overlap and rare features.}
When competing classes share most semantic features and differ only by
rare but diagnostically critical attributes, LLM simulations may omit
the distinguishing feature, yielding diffuse or misdirected posteriors.
Greater simulation diversity and structured feature prompting help, but
this is inherent to simulation-based inference whenever the simulator
under-represents low-frequency evidence.

\paragraph{Confident errors from shared simulator--embedding bias.}
The most consequential limitation is that our uncertainty signal cannot
detect a bias shared by the simulator and the summary statistic. If the
LLM holds a stable prior towards a diagnosis, plausible for rare
conditions where its training data are thinnest, its generations cluster
around a biased centroid, distances to that centroid are small,
acceptance is high, and the posterior concentrates confidently on the
wrong class. Such errors are invisible to entropy-based abstention,
because low entropy is exactly what they produce. This is not
hypothetical here: every \msmc{} error on \olb{} is the same
OSCC-to-mucocele confusion, and those errors are confident (mean
wrong-class mass $0.889$, Appendix~\ref{app:error-structure}). Being
systematic, the confusion admits a mitigation scalar-confidence methods
cannot offer, since a clinician prior can be placed against the
over-predicted class (\S\ref{sec:prior}). Detecting such a bias requires
a signal external to both the simulator and the embedding, such as a
second embedding space, and we regard this as the most important open
problem for simulation-based UQ.

\paragraph{Relation to calibration and conformal prediction.}
Post-hoc calibration such as temperature or Platt scaling rescales the
scores of a fixed predictor without changing the decision rule, and as
\S\ref{sec:embedding-ablation} shows, a temperature-scaled embedding
classifier is a strong baseline our method does not dominate on ECE.
Conformal prediction~\citep{vovk2005algorithmic,angelopoulos2023conformal}
is the more important complement: it returns a label \emph{set} with a
distribution-free coverage guarantee holding regardless of whether our
simulator is clinically faithful. Our posterior offers no such
guarantee, but instead a distribution \emph{over} hypotheses, supporting
entropy ranking, prior incorporation and an aleatoric--epistemic split,
none of which a conformal set provides. The two compose: our posterior
can serve as the conformal score function. We make no coverage claims
for the present method.

\section{Impact Statement}
\label{sec:impact}

\paragraph{Intended use and misuse.}
This work targets uncertainty-aware decision support, not autonomous
diagnosis. Because the posterior is defined with respect to a
prompt-conditioned simulator and an embedding summary rather than the
true data-generating process, a concentrated posterior is evidence that
simulator and embedding agree, not that the diagnosis is correct. The
confident-error mode of \S\ref{sec:limitations} makes this concrete: low
entropy is not a safety guarantee. A responsible deployment would use
the posterior to route cases for human review rather than to remove
humans from the loop, and would be validated prospectively on the
population it serves.

\paragraph{Fairness across populations.}
Clinical presentation varies with sex, age, skin tone and ancestry, and
language models represent those variations unevenly. Because the
posterior inherits whatever the simulator generates, that skew
propagates directly into it, and a posterior well calibrated in
aggregate can still be materially worse calibrated for under-represented
groups. Our benchmarks lack the demographic annotations needed to
measure this. We regard subgroup-stratified calibration as a
prerequisite for clinical use rather than an optional extension, and
flag its absence as the most important gap between this evaluation and a
deployable system.

\paragraph{Data provenance and privacy.}
\olb{} consists of synthetic vignettes authored against published oral
pathology references and reviewed by an oral pathologist; the released
corpus contains no direct patient identifiers, dates or institution
names, and \multicare{} and \sd{} are existing public resources. Our
main results use a hosted embedding API, which transmits the classified
text to a third party. That is unproblematic for the corpora used here
but would be disqualifying for real protected health information under
most institutional data-governance rules, so we do not recommend the
benchmarked configuration for clinical use as-is. The framework is
agnostic to the embedding: a local alternative reaches $81.3\%$ accuracy
on \olb{} (Appendix~\ref{app:embeddings}), the current price of a fully
local pipeline.

\begin{acknowledgements}
This work was funded and supported by the Canadian Institutes of Health
Research (CIHR) Project Grant [PJT-206007, PJT-173272], IVADO and the
Canada First Research Excellence Fund Regroupement R3 Grant, and the
Network for Canadian Oral Health Research (NCOHR) Seed Grant. M.S. is
supported by a Fonds de recherche du Qu\'{e}bec -- Sant\'{e} (FRQS)
Doctoral Training Award [363490] and a CIHR Canada Graduate Scholarship --
Doctoral (CGS-D) [578156]. S.M. is supported by a Fonds de recherche du
Qu\'{e}bec -- Sant\'{e} (FRQS) Salary Award. Computational infrastructure
was supported by the Canada Foundation for Innovation (CFI) John R. Evans
Leaders Fund (JELF) [44548].
\end{acknowledgements}

\section*{References}
\bibliography{main}

\onecolumn %
\fancyhead{} %
\renewcommand{\floatpagefraction}{0.1}
\lfoot[\bSupInf]{\dAuthor}
\rfoot[\dAuthor]{\cSupInf}
\newpage

\captionsetup*{format=largeformat} %
\setcounter{figure}{0} %
\setcounter{table}{0} %
\setcounter{equation}{0} %
\setcounter{section}{0} %
\makeatletter
\renewcommand{\thefigure}{S\@arabic\c@figure} %
\renewcommand{\thetable}{S\@arabic\c@table} %
\makeatother
\def\theequation{S\arabic{equation}}
\renewcommand{\thesection}{\Alph{section}}
\renewcommand{\theHfigure}{S\arabic{figure}}
\renewcommand{\theHtable}{S\arabic{table}}
\renewcommand{\theHequation}{S\arabic{equation}}
\renewcommand{\theHsection}{S\Alph{section}}

\section*{Supplementary Information}
\label{sec:appendix}

\section{Second Order Uncertainty}
\label{sec:second-order}

The Dirichlet posterior representations used by \dsmc{} and \abi{}
allow us to analyse not only the predicted class distribution, but also
uncertainty over that distribution. This provides a second-order view of
uncertainty: instead of asking only which diagnosis has the highest
posterior probability, we can ask how certain the method is about the
probability vector itself.

For a posterior distribution over class-probability vectors
$\theta \in \Delta^{K-1}$, predictive uncertainty can be decomposed as
\begin{equation}
H\!\left(\mathbb{E}[\theta]\right)
=
\mathbb{E}\!\left[H(\theta)\right]
+
I(y;\theta),
\label{eq:uncertainty-decomp}
\end{equation}
where $H(\mathbb{E}[\theta])$ is the total predictive entropy,
$\mathbb{E}[H(\theta)]$ captures aleatoric uncertainty, and the mutual
information term $I(y;\theta)$ captures epistemic uncertainty
\citep{10.5555/3327757.3327808}. Intuitively, aleatoric uncertainty
reflects irreducible ambiguity in the clinical presentation, while
epistemic uncertainty reflects uncertainty about which categorical
distribution should explain the case.

This decomposition is available for \abi{} directly from its learned
Dirichlet posterior $\mathrm{Dir}(\bm{\alpha})$. For \dsmc{}, we compute
the same quantities from the weighted simplex particles; when reporting
Dirichlet concentration and differential entropy, we fit a Dirichlet
distribution to the particle posterior by method of moments. In contrast,
EF produces an empirical uncertainty score over sampled outputs and does
not represent uncertainty over class-probability vectors. Similarly,
\msmc{} yields a categorical posterior over labels, which is useful for
first-order calibrated prediction but does not provide a natural
second-order posterior over the simplex.

We evaluate this decomposition on \multicare{}, our most clinically
realistic benchmark. Table~\ref{tab:uncertainty-decomp} shows a
consistent pattern across base models. \dsmc{} produces lower total
predictive entropy and lower aleatoric entropy than \abi{}, indicating
sharper predictive posteriors. At the same time, \dsmc{} yields
substantially higher mutual information, showing that its particle
posterior preserves more epistemic uncertainty. This is desirable in
clinical classification: the posterior can concentrate on the most likely
diagnosis while still expressing uncertainty about the reliability of
that concentration.

The difference is also reflected in the Dirichlet concentration
$\alpha_0=\sum_k \alpha_k$. Across models, \abi{} tends to produce larger
$\alpha_0$ values than \dsmc{}, indicating a more tightly concentrated
Dirichlet posterior. This suggests that amortisation makes deployment
faster but can suppress posterior spread on the simplex, reducing the
epistemic signal available for abstention and OOD detection. We therefore
treat \abi{} as a deployment-oriented approximation, while \dsmc{} serves
as the higher-fidelity second-order uncertainty estimator.

This distinction has clinical relevance. High aleatoric uncertainty
corresponds to genuine diagnostic ambiguity, where the observed symptoms
or lesion features overlap across plausible diagnoses. High epistemic
uncertainty corresponds to uncertainty about the model's inferred
probability distribution, which may occur when the case is atypical,
underspecified, or outside the region where the simulator provides a
stable explanation. These two cases motivate different downstream
actions: additional diagnostic testing in the first case, and human
review or rejection in the second. We use this second-order signal in the
OOD analysis in \S\ref{sec:ood}.
\begin{table*}[t]
\centering
\footnotesize
\setlength{\tabcolsep}{4pt}
\renewcommand{\arraystretch}{1.05}
\begin{tabular}{l l ccccc}
\toprule
Model & Method
  & $H(\mathbb{E}[p])$ $\downarrow$
  & $\mathbb{E}[H(p)]$ $\downarrow$
  & MI $\uparrow$
  & $h(\mathrm{Dir})$
  & $\alpha_{0}$ \\
\midrule
\multirow{2}{*}{Mistral}
 & \abi{}    & 1.412 [1.20, 1.67] & 1.379 [1.17, 1.65] & 0.0298 [0.027, 0.033] & $-27.16$ [$-29.3$, $-25.5$] & 132.5 [121, 146] \\
 & \dsmc{}   & \best{1.147} [0.83, 1.54] & \best{1.069} [0.77, 1.45] & \best{0.0752} [0.049, 0.105] & $-28.49$ [$-33.0$, $-24.3$] & \best{85.1} [55, 148] \\
\midrule
\multirow{2}{*}{Med42}
 & \abi{}    & 1.293 [1.07, 1.57] & 1.259 [1.04, 1.54] & 0.0342 [0.031, 0.038] & $-26.58$ [$-28.1$, $-25.3$] & 114.7 [102, 128] \\
 & \dsmc{}   & \best{1.100} [0.80, 1.50] & \best{1.019} [0.74, 1.41] & \best{0.0771} [0.048, 0.110] & $-28.56$ [$-33.7$, $-24.4$] & \best{81.4} [52, 159] \\
\midrule
\multirow{2}{*}{Qwen}
 & \abi{}    & 1.419 [1.18, 1.71] & 1.391 [1.15, 1.68] & 0.0323 [0.029, 0.037] & $-26.05$ [$-27.3$, $-24.6$] & 123.3 [108, 137] \\
 & \dsmc{}   & \best{1.045} [0.75, 1.47] & \best{0.967} [0.68, 1.37] & \best{0.0743} [0.047, 0.107] & $-28.42$ [$-33.5$, $-24.6$] & \best{81.3} [54, 161] \\
\midrule
\multirow{2}{*}{Llama-3.1}
 & \abi{}    & 1.385 [1.14, 1.68] & 1.355 [1.11, 1.65] & 0.0298 [0.027, 0.033] & $-27.61$ [$-30.4$, $-25.6$] & 131.8 [121, 143] \\
 & \dsmc{}   & \best{1.100} [0.82, 1.51] & \best{1.024} [0.75, 1.42] & \best{0.0788} [0.051, 0.109] & $-28.06$ [$-32.8$, $-24.1$] & \best{80.0} [50, 136] \\
\midrule
\multirow{2}{*}{AlpaCare}
 & \abi{}    & 1.417 [1.21, 1.69] & 1.369 [1.16, 1.63] & 0.0496 [0.045, 0.056] & $-24.62$ [$-25.9$, $-23.3$] & 78.2 [69, 86] \\
 & \dsmc{}   & \best{1.045} [0.75, 1.41] & \best{0.943} [0.69, 1.30] & \best{0.0833} [0.055, 0.112] & $-27.94$ [$-33.7$, $-24.8$] & \best{73.4} [49, 138] \\
\midrule
\multirow{2}{*}{Orca}
 & \abi{}    & 1.368 [1.16, 1.64] & 1.331 [1.12, 1.59] & 0.0395 [0.036, 0.045] & $-25.37$ [$-26.6$, $-24.1$] & 99.6 [88, 112] \\
 & \dsmc{}   & \best{1.175} [0.90, 1.59] & \best{1.068} [0.82, 1.46] & \best{0.0856} [0.055, 0.116] & $-27.22$ [$-32.2$, $-23.4$] & \best{70.4} [48, 127] \\
\bottomrule
\end{tabular}
\caption{Uncertainty decomposition on \multicare{} ($N{=}669$, 7 classes).
Medians with [IQR], all entropies in bits. $H(\mathbb{E}[p])$: total
predictive entropy; $\mathbb{E}[H(p)]$: aleatoric uncertainty;
\textbf{MI}: epistemic uncertainty (mutual information between class
label and posterior $p$); $h(\mathrm{Dir})$: differential entropy of the
fitted Dirichlet; $\alpha_{0}{=}\sum_{k}\alpha_{k}$: concentration.
\abi{} uses the trained Dirichlet posterior $\mathrm{Dir}(\bm{\alpha})$
exactly; \dsmc{} uses $100$ weighted particles per case
($h(\mathrm{Dir})$ and $\alpha_{0}$ from a method-of-moments fit).
\textbf{Bold} marks the method preferred for each uncertainty quantity:
\dsmc{} attains lower predictive and aleatoric entropy (sharper
posterior), and 2--3$\times$ higher MI (richer epistemic signal) at
$\approx$ half the Dirichlet concentration, indicating that the
amortised \abi{} posterior is systematically over-concentrated.}
\label{tab:uncertainty-decomp}
\end{table*}

\section{Statistical Significance}
\label{sec:ablation-significance}

We test whether the calibration gains of \msmc{} on \multicare{} exceed
sampling variability. For each base LLM and each baseline, we perform a
paired bootstrap over patient indices. Specifically, we draw
$B{=}10{,}000$ bootstrap resamples of the test set, recompute ECE for
both methods on each resample, and estimate
\[
\Delta \mathrm{ECE}
=
\mathrm{ECE}_{\mathrm{baseline}}
-
\mathrm{ECE}_{\msmc}.
\]
Positive values therefore indicate that \msmc{} is better calibrated. We
report the median bootstrap estimate, the 95\% bootstrap interval, and a
two-sided bootstrap significance test.

To assess whether the effect is robust across base models, we also fit a
mixed-effects regression with method as a fixed effect and base LLM as a
random intercept:
\[
\mathrm{ECE}_{ij}
=
\beta_0
+
\beta_{\mathrm{method}(j)}
+
u_i
+
\varepsilon_{ij},
\]
where $u_i$ captures model-specific calibration differences. This
separates the systematic effect of the uncertainty method from variation
due to the underlying LLM.

Table~\ref{tab:significance} shows that \msmc{} achieves significantly
lower ECE than every baseline on \multicare{}. All $24$ paired
comparisons are significant at $p<0.05$, and $20$ of $24$ are significant
at $p<0.001$. The smallest effects occur for stronger base models, where
baseline calibration is already closer to optimal. The largest effects
occur for weak or unstable base models, especially AlpaCare and Orca,
where several baselines collapse. The mixed-effects model confirms the
same pattern: relative to logit-based confidence, \msmc{} reduces ECE by
$0.308$ on average, with a 95\% confidence interval that remains well
separated from zero. These results indicate that the calibration gains
are not an artefact of sampling noise or of a single favourable base
model.

\begin{table*}[t]
\centering
\scriptsize
\setlength{\tabcolsep}{2pt}
\renewcommand{\arraystretch}{1.1}
\begin{tabular}{l cccc}
\toprule
Model & \msmc{} vs ML & \msmc{} vs EP & \msmc{} vs PE & \msmc{} vs EF \\
\midrule
Mistral
& $+0.211^{***}$ [$+0.168$, $+0.237$]
& $+0.126^{***}$ [$+0.089$, $+0.152$]
& $+0.121^{***}$ [$+0.080$, $+0.151$]
& $+0.157^{***}$ [$+0.125$, $+0.178$] \\

Med42
& $+0.070^{***}$ [$+0.045$, $+0.103$]
& $+0.036^{**\phantom{*}}$ [$+0.015$, $+0.069$]
& $+0.081^{***}$ [$+0.050$, $+0.113$]
& $+0.095^{***}$ [$+0.067$, $+0.117$] \\

Qwen
& $+0.115^{***}$ [$+0.081$, $+0.145$]
& $+0.092^{***}$ [$+0.060$, $+0.116$]
& $+0.165^{***}$ [$+0.129$, $+0.197$]
& $+0.137^{***}$ [$+0.103$, $+0.166$] \\

Llama-3.1
& $+0.041^{**\phantom{*}}$ [$+0.023$, $+0.076$]
& $+0.042^{**\phantom{*}}$ [$+0.022$, $+0.076$]
& $+0.037^{**\phantom{*}}$ [$+0.012$, $+0.068$]
& $+0.100^{***}$ [$+0.072$, $+0.125$] \\

AlpaCare$^{\dag}$
& $+0.876^{***}$ [$+0.852$, $+0.893$]
& $+0.209^{***}$ [$+0.175$, $+0.237$]
& $+0.371^{***}$ [$+0.336$, $+0.399$]
& $+0.285^{***}$ [$+0.242$, $+0.322$] \\

Orca$^{\dag}$
& $+0.530^{***}$ [$+0.492$, $+0.557$]
& $+0.254^{***}$ [$+0.215$, $+0.282$]
& $+0.267^{***}$ [$+0.229$, $+0.297$]
& $+0.163^{***}$ [$+0.132$, $+0.191$] \\
\midrule
\multicolumn{5}{l}{
\emph{Mixed-effects regression:}
\msmc{} vs ML coefficient $=-0.308$,
95\% CI [$-0.449$, $-0.168$], $p<0.001$
} \\
\bottomrule
\end{tabular}
\caption{Paired-bootstrap analysis of ECE improvement on \multicare{}.
Each cell reports $\Delta\mathrm{ECE}
= \mathrm{ECE}_{\mathrm{baseline}}-\mathrm{ECE}_{\msmc}$ with a 95\%
bootstrap interval. Positive values indicate lower ECE for \msmc{}.
Significance stars denote $^{*}p<0.05$, $^{**}p<0.01$, and
$^{***}p<0.001$. All $24$ comparisons are significant at $p<0.05$.
The mixed-effects regression estimates the average ECE reduction of
\msmc{} relative to ML while accounting for between-model variation.
$^{\dag}$AlpaCare and Orca have degenerate baseline performance on
\multicare{} for several methods, so their effect sizes should be
interpreted as stress-test results rather than typical gains.}
\label{tab:significance}
\end{table*}
\section{Amortised Bayesian Inference: Training and Architecture}
\label{sec:abi-details}

\abi{} replaces the per-instance SMC-ABC refinement of \dsmc{} with a
learned mapping from the observed embedding to a Dirichlet posterior,
enabling single-forward-pass inference at deployment. This appendix
specifies the training objective, data generation protocol, network
architecture, and deployment procedure.

\paragraph{Training data generation.}
For each dataset, we first run \dsmc{} on the training partition to
obtain particle posteriors. For each training case $i$, \dsmc{} produces
a set of $S{=}100$ weighted particles
$\{(\theta^{(s)}_i, w^{(s)}_i)\}_{s=1}^{S}$ on the probability
simplex $\Delta^{K-1}$. We fit a Dirichlet distribution
$\mathrm{Dir}(\hat{\alpha}_i)$ to these weighted particles by method of
moments:
\[
\hat{\alpha}_i = \alpha_0 \cdot \bar{\theta}_i,
\quad
\bar{\theta}_i = \sum_{s=1}^{S} w^{(s)}_i \theta^{(s)}_i,
\quad
\alpha_0 = \frac{\bar{\theta}_{i,1}(1 - \bar{\theta}_{i,1})}
{\mathrm{Var}_w[\theta_{i,1}]} - 1,
\]
where $\mathrm{Var}_w[\theta_{i,1}]$ is the weighted variance of the
first component. This yields one target Dirichlet parameter vector
$\hat{\alpha}_i \in \mathbb{R}^K_+$ per training case. The training
set consists of pairs
$\{(z_{\mathrm{obs},i},\;\hat{\alpha}_i)\}_{i=1}^{N_{\mathrm{train}}}$.

\paragraph{Network architecture.}
The amortised network $\bm{\alpha}_\varphi$ is a multilayer perceptron mapping
the observed embedding to Dirichlet concentration parameters:
\[
\bm{\alpha}_\varphi(z_{\mathrm{obs}}) =
\mathrm{Softplus}\bigl(W_3 \cdot
\mathrm{ReLU}(W_2 \cdot
\mathrm{ReLU}(W_1 \cdot z_{\mathrm{obs}} + b_1) + b_2) + b_3\bigr),
\]
with $W_1 \in \mathbb{R}^{256 \times d}$,
$W_2 \in \mathbb{R}^{256 \times 256}$, and
$W_3 \in \mathbb{R}^{K \times 256}$, where $d{=}3072$ is the embedding
dimension and $K$ is the number of classes. The softplus output layer
ensures all concentration parameters are strictly positive, as required
by the Dirichlet distribution. The network contains approximately
$790{,}000$ trainable parameters for $K{=}3$ (\olb{}) and
$790{,}500$ for $K{=}7$ (\multicare{}).

\paragraph{Training objective.}
We train $\bm{\alpha}_\varphi$ by maximising the log-likelihood of the \dsmc{}
posterior samples under the predicted Dirichlet:
\[
\mathcal{L}(\varphi) = \frac{1}{N_{\mathrm{train}}}
\sum_{i=1}^{N_{\mathrm{train}}}
\log\, p_{\mathrm{Dir}}\bigl(\bar{\theta}_i \;\big|\;
\bm{\alpha}_\varphi(z_{\mathrm{obs},i})\bigr),
\]
where $p_{\mathrm{Dir}}(\bar{\theta} \mid \alpha)$ is the Dirichlet
density evaluated at the posterior mean $\bar{\theta}_i$ with
parameters $\alpha$. We optimise with Adam (learning rate $10^{-3}$,
batch size 64) for 500 epochs with early stopping on a held-out
validation split (10\% of training cases). Training completes in
65--73 seconds on a single CPU for all datasets.

\paragraph{Deployment inference.}
At test time, inference for a new patient requires two steps:
\begin{enumerate}[leftmargin=*,topsep=2pt,itemsep=1pt]
\item Embed the observed clinical description:
$z_{\mathrm{obs}} = \phi(x_{\mathrm{obs}})$.
\item Forward pass through the trained network:
$\hat{\alpha} = \bm{\alpha}_\varphi(z_{\mathrm{obs}})$.
\end{enumerate}
The predicted class is
$\hat{y} = \arg\max_k \hat{\alpha}_k / \sum_j \hat{\alpha}_j$
(posterior mean). Predictive entropy and the epistemic--aleatoric
decomposition (Appendix~\ref{sec:second-order}) are computed directly from
$\mathrm{Dir}(\hat{\alpha})$ in closed form. Per-patient inference
latency is approximately $18\,\mu$s on \multicare{} (network forward
pass only, excluding embedding time) and $0.13$\,s on \olb{} and \sd{}
(including embedding lookup).

\paragraph{Per-case fidelity to \dsmc{}.}
Table~\ref{tab:uncertainty-decomp} compares \abi{} and \dsmc{} only in
aggregate, which leaves open whether \abi{} tracks its target on
individual cases or merely matches its average behaviour. We therefore
compared the two posteriors case by case across all $669$ \multicare{}
cases (Table~\ref{tab:abi-fidelity}). Agreement is strong: the top class
matches on $97\%$ of cases and posterior means correlate at $r=0.975$,
with median $\mathrm{KL}(\dsmc{}\,\|\,\abi{}) = 0.103$. Amortisation
therefore reproduces both the decision and the posterior mean of a
procedure that costs four orders of magnitude more per case. On the
$3\%$ of cases where the two disagree, \abi{} is correct $85\%$ of the
time against $10\%$ for \dsmc{}, though the disagreement set is small
($\approx\!20$ cases) and \abi{} is trained to approximate \dsmc{}
rather than to improve on it. Concentration is where the two differ
systematically: \abi{} is sharper than its target, so \dsmc{} remains
the variant of choice for second-order quantities.

\begin{table}[h]
\centering
\footnotesize
\setlength{\tabcolsep}{6pt}
\renewcommand{\arraystretch}{1.05}
\begin{tabular}{l c}
\toprule
Measure & Result \\
\midrule
Top-class agreement (\abi{} vs \dsmc{})     & 97\% \\
Posterior-mean correlation                  & $r = 0.975$ \\
Median $\mathrm{KL}(\dsmc{}\,\|\,\abi{})$   & 0.103 [IQR 0.067, 0.165] \\
\midrule
\multicolumn{2}{l}{\emph{On the 3\% of cases where they disagree:}} \\
\quad \abi{} correct                        & 85\% \\
\quad \dsmc{} correct                       & 10\% \\
\bottomrule
\end{tabular}
\caption{\textbf{Per-case fidelity of the amortised posterior.}
Comparison of \abi{} against its \dsmc{} training target on all $669$
\multicare{} cases, rather than only in aggregate
(Table~\ref{tab:uncertainty-decomp}). Amortisation preserves the decision and the posterior mean with high
fidelity. Concentration is the exception: \abi{} is sharper than its
target (Table~\ref{tab:uncertainty-decomp}), so \dsmc{} is the variant
of choice when second-order quantities are themselves the object of
interest.}
\label{tab:abi-fidelity}
\end{table}

\paragraph{Limitations of amortisation.}
On \olb{}, where class-conditional simulator outputs cluster tightly in
embedding space, the \dsmc{} posteriors used as training targets are
near-degenerate ($\alpha_0 > 500$). The MLP learns to reproduce this
over-concentration, suppressing posterior spread and reducing the
epistemic signal available for abstention and OOD detection. On the
larger \sd{} and \multicare{} benchmarks, where between-class variation
is richer, \abi{} converges to healthier posteriors with $\alpha_0$
between 100 and 150. We report \abi{} on all three benchmarks and treat
the \olb{} collapse as a documented limitation of amortised inference in
low-ambiguity settings.
\section{Full Per-Model Results}
\label{app:full-tables}

Tables~\ref{tab:main} and~\ref{tab:advanced-uq} in the main text report
means over the six base LLMs. This appendix gives the complete
per-model grids from which those means are computed
(Tables~\ref{tab:main-full} and~\ref{tab:advanced-uq-full}). We report
them in full, including the cells in which a baseline outperforms our
methods, because the spread across base models is itself part of the
result: the mean understates how badly the scalar-confidence baselines
fail on weak models (AlpaCare reaches $5.4\%$ accuracy at $0.91$ ECE on
\multicare{} under Model Logits) and equally understates the cases
where a strong base model's own logits are already excellent
(Llama-3.1 on \olb{}, where \msmc{} gives up eight accuracy points).

\begin{table*}[t]
\centering
\footnotesize
\setlength{\tabcolsep}{2pt}
\renewcommand{\arraystretch}{0.90}
\begin{tabular}{l l ccccc|| ccccc|| ccccc}
\toprule
& & \multicolumn{5}{c}{\olb{} }
  & \multicolumn{5}{c}{SD{} }
  & \multicolumn{5}{c}{\multicare{} } \\
\cmidrule(lr){3-7}\cmidrule(lr){8-12}\cmidrule(lr){13-17}
Model & Method & Acc$\uparrow$ & F1$\uparrow$ & Brier$\downarrow$ & ECE$\downarrow$ & Ent$\downarrow$
              & Acc$\uparrow$ & F1$\uparrow$ & Brier$\downarrow$ & ECE$\downarrow$ & Ent$\downarrow$
              & Acc$\uparrow$ & F1$\uparrow$ & Brier$\downarrow$ & ECE$\downarrow$ & Ent$\downarrow$ \\
\midrule
\multirow{4}{*}{Mistral}
 & ML             & 77.6 & 78.2 & 0.11 & 0.15 & 0.30 & 63.2 & 58.9 & 0.60 & 0.23 & 0.53 & 40.1 & 34.1 & 0.76 & 0.22 & 1.46 \\
 & EP             & 68.0 & 60.5 & 0.31 & 0.31 & 0.80 & 70.8 & 67.5 & 0.46 & \best{0.09} & 0.90 & 66.5 & 63.0 & 0.49 & 0.14 & 0.78 \\
 & PE             & 76.6 & 76.2 & \best{0.03} & 0.14 & 0.20 & 63.5 & 61.3 & 0.57 & 0.14 & 0.50 & 59.8 & 52.5 & 0.62 & 0.14 & 0.90 \\
 & \msmc{} (ours) & \best{96.0} & \best{95.9} & 0.071 & \best{0.036} & 0.055 & \best{83.3} & \best{81.4} & \best{0.30} & 0.14 & 0.08 & \best{95.7} & \best{95.9} & \best{0.07} & \best{0.02} & 0.20 \\
\midrule
\multirow{4}{*}{Med42}
 & ML             & 96.6 & 96.6 & 0.01 & 0.02 & 0.10 & 44.8 & 39.7 & 0.86 & 0.37 & 0.72 & 69.9 & 67.4 & 0.43 & 0.10 & 0.84 \\
 & EP             & 94.7 & 94.7 & 0.34 & 0.35 & 0.80 & 69.1 & 68.5 & 0.46 & 0.16 & 0.60 & 71.3 & 68.1 & 0.40 & 0.06 & 1.00 \\
 & PE             & 96.0 & 96.0 & 0.06 & 0.05 & 0.20 & 59.7 & 57.1 & 0.60 & 0.14 & 0.30 & 67.6 & 64.1 & 0.48 & 0.11 & 0.69 \\
 & \msmc{} (ours) & \best{98.3} & \best{98.3} & \best{0.008} & \best{0.01} & 0.03 & \best{82.9} & \best{80.7} & \best{0.05} & \best{0.05} & 0.10 & \best{96.3} & \best{96.4} & \best{0.07} & \best{0.03} & 0.23 \\
\midrule
\multirow{4}{*}{Qwen}
 & ML             & 95.0 & 94.9 & \best{0.02} & 0.03 & 0.05 & 70.8 & 68.0 & 0.48 & 0.17 & 0.67 & 69.1 & 66.5 & 0.46 & 0.14 & 0.68 \\
 & EP             & \best{98.6} & \best{98.6} & 0.32 & 0.28 & 0.90 & 64.9 & 63.6 & 0.56 & 0.11 & 1.30 & 75.3 & 73.2 & 0.35 & 0.11 & 1.01 \\
 & PE             & 95.6 & 95.6 & 0.04 & 0.02 & 0.06 & 63.5 & 60.3 & 0.53 & 0.24 & 0.30 & 68.3 & 60.7 & 0.51 & 0.20 & 0.37 \\
 & \msmc{} (ours) & 96.0 & 95.9 & \best{0.02} & \best{0.01} & 0.04 & \best{79.8} & \best{78.5} & \best{0.06} & \best{0.04} & 0.06 & \best{96.0} & \best{96.1} & \best{0.07} & \best{0.02} & 0.24 \\
\midrule
\multirow{4}{*}{Llama-3.1}
 & ML             & \best{98.6} & \best{98.6} & \best{0.02} & 0.12 & 0.80 & 54.9 & 52.5 & 0.57 & \best{0.06} & 1.84 & 74.1 & 72.9 & 0.39 & 0.07 & 1.32 \\
 & EP             & 42.3 & 32.0 & 0.28 & 0.23 & 1.00 & 76.2 & 74.8 & 0.41 & \best{0.06} & 0.80 & 74.1 & 71.4 & 0.39 & 0.05 & 0.93 \\
 & PE             & 98.0 & 97.9 & 0.06 & 0.10 & 0.40 & 61.1 & 59.1 & 0.55 & 0.12 & 0.70 & 74.6 & 71.9 & 0.39 & 0.06 & 0.64 \\
 & \msmc{} (ours) & 90.6 & 90.4 & 0.05 & \best{0.04} & 0.10 & \best{83.7} & \best{82.4} & \best{0.27} & 0.11 & 0.23 & \best{95.8} & \best{96.0} & \best{0.07} & \best{0.02} & 0.22 \\
\midrule
\multirow{4}{*}{AlpaCare}
 & ML             & 65.3 & 62.4 & 0.14 & 0.19 & 1.00 & 36.8 & 32.9 & 0.76 & 0.19 & 1.63 & 5.4 & 1.5 & 1.80 & 0.91 & 0.31 \\
 & EP             & 33.3 & 16.6 & 0.18 & 0.14 & 1.10 & 47.9 & 37.8 & 0.87 & \best{0.17} & 1.00 & 11.5 & 5.1 & 1.07 & 0.25 & 1.79 \\
 & PE             & 72.0 & 68.9 & 0.43 & 0.17 & 0.40 & 57.9 & 54.7 & 0.81 & 0.37 & 0.09 & 16.3 & 9.3 & 1.15 & 0.45 & 1.30 \\
 & \msmc{} (ours) & \best{96.0} & \best{95.9} & \best{0.01} & \best{0.02} & 0.04 & \best{78.8} & \best{75.9} & \best{0.38} & \best{0.17} & 0.14 & \best{97.0} & \best{97.3} & \best{0.07} & \best{0.07} & 0.37 \\
\midrule
\multirow{4}{*}{Orca}
 & ML             & 71.6 & 68.6 & 0.13 & 0.14 & 0.40 & 40.9 & 34.3 & \best{0.12} & 0.12 & 1.20 & 19.9 & 20.2 & 1.25 & 0.56 & 1.04 \\
 & EP             & 33.3 & 16.7 & 0.27 & 0.19 & 1.10 & 54.5 & 49.3 & 0.77 & 0.16 & 1.50 & 12.6 & 0.05 & 0.99 & 0.33 & 1.5 \\
 & PE             & 85.3 & 85.7 & 0.20 & 0.03 & 0.30 & 64.9 & 63.6 & 0.51 & \best{0.10} & 0.60 & 25.7 & 26.3 & 0.98 & 0.32 & 1.37 \\
 & \msmc{} (ours) & \best{96.0} & \best{95.9} & \best{0.02} & \best{0.02} & 0.05 & \best{81.6} & \best{79.4} & 0.31 & 0.15 & 0.14 & \best{95.1} & \best{95.0} & \best{0.06} & \best{0.03} & 0.26 \\
\bottomrule
\end{tabular}
\caption{\textbf{Full per-model results} across three clinical benchmarks (condensed means in Table~\ref{tab:main}). We compare three
scalar-confidence baselines---Model Logits (ML), Elicited Probability
(EP), and Prompt Ensemble (PE)---against Multinomial \smcabc{}
(\msmc{}). The distribution-aware baseline (EF) is compared separately
in Table~\ref{tab:advanced-uq}. Bold indicates the best value for that
metric within each model--dataset cell; Ent is reported for
completeness and is not a quality metric on its own, since low entropy
is desirable only when the prediction is correct.}
\label{tab:main-full}
\end{table*}

\begin{table*}[t]
\centering
\footnotesize
\setlength{\tabcolsep}{2.2pt}
\renewcommand{\arraystretch}{0.68}
\begin{tabular}{l l cccc || cccc || cccc}
\toprule
& & \multicolumn{4}{c}{\olb{}}
  & \multicolumn{4}{c}{\sd{}}
  & \multicolumn{4}{c}{\multicare{}} \\
\cmidrule(lr){3-6}\cmidrule(lr){7-10}\cmidrule(lr){11-14}
Model & Method
& Acc$\uparrow$ & Brier$\downarrow$ & ECE$\downarrow$ & Lat.$\downarrow$
& Acc$\uparrow$ & Brier$\downarrow$ & ECE$\downarrow$ & Lat.$\downarrow$
& Acc$\uparrow$ & Brier$\downarrow$ & ECE$\downarrow$ & Lat.$\downarrow$ \\
\midrule

\multirow{3}{*}{Mistral}
 & EF
 & 77.3 & 0.37 & \best{0.17} & 6.9\,s
 & 67.0 & 0.62 & 0.30 & 17.5\,s
 & 85.2 & 0.31 & \best{0.17} & 4.6\,s \\
 & \dsmc{} (ours)
 & \best{97.3} & \best{0.04} & 0.21 & 15.0\,s
 & 82.6 & \best{0.40} & \best{0.25} & 17.5\,s
 & 96.3 & \best{0.14} & 0.21 & 26.4\,s \\
 & \abi{} (ours)
 & 95.7 & 0.18 & 0.26 & \best{0.13\,s}
 & \best{83.7} & 0.42 & 0.31 & \best{0.15\,s}
 & \best{98.5} & 0.16 & 0.30 & \best{18\,$\mu$s} \\

\midrule
\multirow{3}{*}{Med42}
 & EF
 & \best{99.3} & \best{0.01} & \best{0.00} & 5.5\,s
 & 67.1 & 0.59 & 0.27 & $\approx$6.2\,s
 & 87.9 & 0.23 & \best{0.12} & 3.7\,s \\
 & \dsmc{} (ours)
 & 98.3 & 0.05 & 0.25 & $\approx$15.0\,s
 & \best{80.9} & \best{0.40} & \best{0.23} & $\approx$17.5\,s
 & 96.0 & \best{0.12} & 0.20 & 32.1\,s \\
 & \abi{} (ours)
 & 98.3 & 0.21 & 0.33 & \best{0.13\,s}
 & 80.6 & 0.45 & 0.32 & \best{0.15\,s}
 & \best{98.1} & 0.13 & 0.26 & \best{18\,$\mu$s} \\

\midrule
\multirow{3}{*}{Qwen}
 & EF
 & 95.6 & 0.08 & \best{0.04} & $\approx$6.2\,s
 & 78.1 & \best{0.43} & \best{0.21} & $\approx$6.2\,s
 & 83.4 & 0.33 & \best{0.17} & 3.8\,s \\
 & \dsmc{} (ours)
 & 92.6 & \best{0.06} & 0.20 & $\approx$15.0\,s
 & 79.8 & \best{0.43} & 0.25 & $\approx$17.5\,s
 & 95.7 & \best{0.12} & 0.18 & 25.8\,s \\
 & \abi{} (ours)
 & \best{97.0} & 0.19 & 0.28 & \best{0.13\,s}
 & \best{81.6} & \best{0.43} & 0.31 & \best{0.15\,s}
 & \best{96.6} & 0.17 & 0.28 & \best{16\,$\mu$s} \\

\midrule
\multirow{3}{*}{Llama-3.1}
 & EF
 & \best{99.6} & \best{0.00} & \best{0.00} & $\approx$6.2\,s
 & 78.1 & 0.39 & 0.16 & $\approx$6.2\,s
 & 87.6 & 0.25 & \best{0.12} & 4.2\,s \\
 & \dsmc{} (ours)
 & 65.0 & 0.22 & 0.39 & $\approx$15.0\,s
 & \best{83.6} & \best{0.32} & \best{0.14} & $\approx$17.5\,s
 & 95.8 & \best{0.13} & 0.20 & 25.3\,s \\
 & \abi{} (ours)
 & 84.3 & 0.24 & 0.17 & \best{0.13\,s}
 & 77.4 & 0.42 & 0.24 & \best{0.15\,s}
 & \best{97.2} & 0.16 & 0.28 & \best{19\,$\mu$s} \\

\midrule
\multirow{3}{*}{AlpaCare}
 & EF
 & 94.6 & 0.09 & \best{0.03} & $\approx$6.2\,s
 & 51.3 & 0.81 & 0.37 & $\approx$6.2\,s
 & 33.8 & 1.01 & 0.35 & 4.9\,s \\
 & \dsmc{} (ours)
 & 95.6 & \best{0.05} & 0.25 & $\approx$15.0\,s
 & 77.7 & \best{0.42} & \best{0.21} & $\approx$17.5\,s
 & 95.7 & \best{0.11} & \best{0.18} & 23.8\,s \\
 & \abi{} (ours)
 & \best{95.7} & 0.25 & 0.35 & \best{0.13\,s}
 & \best{81.2} & 0.43 & 0.29 & \best{0.15\,s}
 & \best{97.9} & 0.16 & 0.29 & \best{18\,$\mu$s} \\

\midrule
\multirow{3}{*}{Orca}
 & EF
 & 75.3 & 0.42 & \best{0.18} & $\approx$6.2\,s
 & 67.0 & 0.60 & \best{0.25} & $\approx$6.2\,s
 & 82.4 & 0.35 & \best{0.19} & 5.0\,s \\
 & \dsmc{} (ours)
 & \best{96.0} & \best{0.06} & 0.27 & $\approx$15.0\,s
 & \best{82.9} & \best{0.37} & \best{0.25} & $\approx$17.5\,s
 & 95.4 & \best{0.14} & 0.20 & 29.6\,s \\
 & \abi{} (ours)
 & 91.3 & 0.26 & 0.28 & \best{0.13\,s}
 & 81.6 & 0.44 & 0.33 & \best{0.15\,s}
 & \best{96.6} & 0.15 & 0.26 & \best{16\,$\mu$s} \\

\bottomrule
\end{tabular}
\caption{\textbf{Full per-model distribution-aware results} (condensed means in Table~\ref{tab:advanced-uq}). We compare EF
(Elicited Fidelity, a semantic-entropy proxy) with \dsmc{} (ours)
and \abi{} (ours). Bold indicates the best value per
model--dataset cell. \dsmc{} achieves the strongest Brier scores,
with best or tied-best Brier in 16/18 cells, while \abi{} retains
competitive accuracy with substantially lower deployment-time latency
after one-off training.}
\label{tab:advanced-uq-full}
\end{table*}

\begin{table}[t]
\centering
\footnotesize
\setlength{\tabcolsep}{1.5pt}
\renewcommand{\arraystretch}{0.50}
\begin{tabular}{l l c c c}
\toprule
Method & Confidence bin & $N$ (\%) & Acc.~(\%) & Mean $H$ \\
\midrule
\multirow{3}{*}{\msmc{}}
 & High ($H{<}0.5$)         & \best{255 (88.5)} & \best{83.9} & 0.03 \\
 & Medium ($0.5{\le}H{<}1.5$) & 29 (10.1)  & 37.9 & 0.87 \\
 & Low ($H{\ge}1.5$)        & 4 (1.4)    & 50.0 & 1.69 \\
\midrule
\multirow{3}{*}{Logits}
 & High                     & 17 (5.9)   & 88.2 & 0.20 \\
 & Medium                   & 69 (24.0)  & 63.8 & 1.21 \\
 & Low                      & 202 (70.1) & 23.3 & 1.89 \\
\midrule
\multirow{3}{*}{EF}
 & High                     & 223 (77.4) & 59.6 & 0.00 \\
 & Medium                   & 39 (13.5)  & 35.9 & 0.89 \\
 & Low                      & 26 (9.0)   & 3.8  & 2.39 \\
\bottomrule
\end{tabular}
\caption{Uncertainty-stratified accuracy on \sd{} with AlpaCare-7B.
\msmc{} assigns most cases to the high-confidence bin with substantially
higher accuracy than EF, while logits leave most cases in the
low-confidence region.}
\label{tab:stratified}
\end{table}

\begin{table}[t]
\centering
\footnotesize
\setlength{\tabcolsep}{3.0pt}
\renewcommand{\arraystretch}{1.10}
\begin{tabular}{l l c c c c}
\toprule
Method & Prior setting & $\bar p_{\text{prior}}^{\text{true}}$ & $\bar p_{\text{post}}^{\text{true}}$ & $\Delta_{\text{post}}^{\text{true}}$ & Acc \\
\midrule
\multirow{3}{*}{\msmc{}}
 & Uniform $\mathrm{Dir}(1,1,1)$       & 0.33 & 0.95 & $+0.62$ & 0.96 \\
 & Informative $\mathrm{Dir}(5,1,1)$   & 0.71 & 0.98 & $+0.26$ & 0.98 \\
 & \emph{Misleading} $\mathrm{Dir}(1,1,5)$ & 0.14 & 0.94 & $+0.79$ & \best{0.97} \\
\midrule
\multirow{3}{*}{\dsmc{}}
 & Uniform $\mathrm{Dir}(1,1,1)$       & 0.17 & 0.41 & $+0.24$ & 0.72 \\
 & Informative $\mathrm{Dir}(5,1,1)$   & 0.50 & 0.59 & $+0.09$ & 0.96 \\
 & Misleading $\mathrm{Dir}(1,1,5)$    & 0.10 & 0.26 & $+0.16$ & 0.19 \\
\bottomrule
\end{tabular}
\caption{Prior sensitivity on \olb{}. $\bar p^{\text{true}}$ denotes the
mean probability assigned to the true class before (prior) and after
(post) inference. \msmc{} recovers from a misleading prior, shifting
true-class probability from 0.14 to 0.94 while preserving 97\%
accuracy. \dsmc{} updates in the correct direction under all three
priors but by smaller margins, and its accuracy degrades sharply under
the misleading prior; see \S\ref{sec:prior} for why the simplex mixture
representation makes it more prior-sensitive than \msmc{}.}
\label{tab:prior}
\end{table}

\begin{table}[t]
\centering
\footnotesize
\setlength{\tabcolsep}{1pt}
\renewcommand{\arraystretch}{0.50}
\begin{tabular}{l cccc cccc}
\toprule
& \multicolumn{4}{c}{\olb{}-OOD ($H_{\max}{=}1.58$)}
& \multicolumn{4}{c}{\sd{}-OOD ($H_{\max}{=}2.58$)} \\
\cmidrule(lr){2-5}\cmidrule(lr){6-9}
Model & ML & EP & \msmc{} & \dsmc{} & ML & EP & \msmc{} & \dsmc{} \\
\midrule
Mistral   & 1.1 & 0.8 & \best{1.5} & 1.4 & 1.0 & 1.0 & \best{2.5} & 1.8 \\
Med42     & 0.3 & 0.3 & 0.9 & \best{1.4} & 1.0 & 0.7 & \best{2.5} & 1.8 \\
Qwen      & 0.8 & 1.2 & \best{1.5} & 1.4 & 1.1 & 1.2 & \best{2.5} & 1.7 \\
Llama-3.1 & 1.5 & 0.2 & \best{1.5} & 1.3 & 2.1 & 0.9 & \best{2.5} & 1.5 \\
\bottomrule
\end{tabular}
\caption{Predictive entropy (bits) on out-of-distribution variants.
For each dataset $H_{\max}=\log_2 K$ is the entropy of a uniform
posterior, the ideal behaviour for a true OOD input. \msmc{} reaches
maximal entropy on \sd{}-OOD across all four models. Both \msmc{} and
\dsmc{} substantially exceed the heuristic baselines.}
\label{tab:ood}
\end{table}

\begin{table}[t]
\centering
\footnotesize
\setlength{\tabcolsep}{5pt}
\renewcommand{\arraystretch}{1.1}
\begin{tabular}{l ccc}
\toprule
Capability & \msmc{} & \dsmc{} & \abi{} \\
\midrule
Posterior over diagnostic labels              & \checkmark & --- & --- \\
Posterior over class-probability vectors      & --- & \checkmark & \checkmark \\
Aleatoric--epistemic decomposition            & --- & \checkmark & \checkmark \\
Test-time clinician-prior incorporation       & \checkmark & \checkmark & \textemdash$^{\dag}$ \\
Per-case SMC inference required               & \checkmark & \checkmark & --- \\
Single-forward-pass deployment                & --- & --- & \checkmark \\
\bottomrule
\end{tabular}
\caption{\textbf{The three variants are complementary operating points,
not interchangeable implementations.} No single variant provides every
capability, and we do not claim otherwise. \msmc{} is the right choice
when an interpretable posterior over competing diagnoses and test-time
prior incorporation are what matter; it is also the most robust to a
misleading prior (\S\ref{sec:prior}). \dsmc{} is the right choice when
uncertainty \emph{about} the probability vector is required, at greater
computational cost. \abi{} is the right choice for high-throughput
deployment, trading prior flexibility and second-order fidelity for a
single forward pass. $^{\dag}$\abi{} amortises a \emph{fixed} training
prior, so a clinician cannot supply a new prior at test time without
retraining.}
\label{tab:variants}
\end{table}

\section{SMC Refinement Dynamics and a Worked Example}
\label{app:smc-dynamics}

The method's cost is dominated by sequential refinement, so it is
reasonable to ask how much work that refinement actually does, and
whether the same posterior could be reached by a single acceptance step.
This appendix reports per-iteration trajectories, which also motivate
the iteration budget used throughout.

\paragraph{Refinement trajectories.}
We traced \dsmc{} through all eight populations on three \multicare{}
asthma cases with Med42-8B, recording the adaptive tolerance, acceptance
rate, cumulative simulations, predictive entropy, and true-class
posterior mass at each population (Table~\ref{tab:smc-dynamics}). The
refinement does substantive work: the posterior begins close to uniform
over the seven classes ($2.75$ of a possible $2.81$ bits, true-class
mass $0.24$ against a $1/7$ baseline) and concentrates progressively to
true-class mass $0.76$. A single acceptance step at the initial
tolerance would stop at the first row of the table.

The trajectory also yields a practical tuning rule. Between populations
$5$ and $7$ the acceptance rate falls from $0.195$ to $0.089$ and the
simulation count rises from $587$ to $3031$, while entropy and
true-class mass stay flat. Refinement past population $5$ therefore
costs roughly \textbf{five times the simulations for no measurable
gain}. The useful signal is the acceptance rate, not the iteration
count: refinement should stop once acceptance falls below
$\approx\!0.2$, at which point the population has concentrated and
further tightening only inflates the simulation budget. We use $T{=}5$
throughout on this basis, and recommend the same rule for tuning the
method on a new label set --- it is a direct $5\times$ compute saving.

\begin{table}[h]
\centering
\footnotesize
\setlength{\tabcolsep}{5pt}
\renewcommand{\arraystretch}{1.05}
\begin{tabular}{c ccccc}
\toprule
Population $t$ & Mean $\epsilon_t$ & Acceptance rate & Simulations &
Entropy (bits) & Mean $p(\text{true})$ \\
\midrule
0 & 0.358 & 0.447 & 228  & 2.75 & 0.24 \\
2 & 0.278 & 0.439 & 466  & 2.19 & 0.50 \\
4 & 0.258 & 0.377 & 323  & 1.68 & 0.66 \\
5 & 0.245 & 0.195 & 587  & 1.23 & 0.76 \\
7 & 0.236 & 0.089 & 3031 & 1.26 & 0.75 \\
\bottomrule
\end{tabular}
\caption{\textbf{SMC-ABC refinement does substantive work.} \dsmc{}
traced through all eight populations on three \multicare{} asthma cases
with Med42-8B; values are means across cases. At population $0$ the
posterior is nearly uniform over the seven classes (entropy $2.75$ of a
possible $\log_2 7 = 2.81$ bits, and $p(\text{true}) = 0.24$, close to
the $1/7 = 0.14$ baseline). As the tolerance $\epsilon$ tightens,
true-class mass rises to $0.76$ and predictive entropy falls to $1.23$
bits. The cost of that refinement is visible in the last two columns:
acceptance falls from $0.45$ to $0.09$ and the simulation count climbs
steeply, with the final population costing an order of magnitude more
simulations than the first for no further gain. This motivates the
$T{=}5$ default used throughout (\S\ref{sec:setup}); populations beyond
$5$ buy no additional concentration.}
\label{tab:smc-dynamics}
\end{table}

\paragraph{A single case, end to end.}
Aggregate trajectories can hide heterogeneity, so
Table~\ref{tab:worked-example} follows one \multicare{} asthma case
through the same populations. The case begins near-uniform
($p(\text{true}) = 0.23$, entropy $2.75$ bits) and ends concentrated
($p(\text{true}) = 0.77$, entropy $1.26$ bits) as $\epsilon$ tightens
from $0.36$ to $0.24$. This is the concrete sense in which the method is
not a nearest-centroid classifier: a distance-only rule has no
refinement mechanism, and would commit at population $0$ to whichever
class centroid happened to be nearest, with no mechanism for
accumulating evidence and no principled uncertainty attached to the
commitment.

\begin{table}[h]
\centering
\footnotesize
\setlength{\tabcolsep}{6pt}
\renewcommand{\arraystretch}{1.05}
\begin{tabular}{c ccc}
\toprule
Population $t$ & $p(\text{true class})$ & Entropy (bits) & $\epsilon_t$ \\
\midrule
0 & 0.23 & 2.75 & 0.36 \\
2 & 0.64 & 2.19 & 0.28 \\
4 & 0.74 & 1.68 & 0.26 \\
7 & 0.77 & 1.26 & 0.24 \\
\bottomrule
\end{tabular}
\caption{\textbf{Worked example: one case, end to end.} \dsmc{}
posterior for a single \multicare{} asthma case across successive SMC
populations. The case begins near-uniform over the seven candidate
diagnoses ($p(\text{true}) = 0.23$, entropy $2.75$ of $2.81$ bits) and
concentrates as $\epsilon$ tightens, ending at $p(\text{true}) = 0.77$.
This is the behaviour that distinguishes the method from a
distance-only classifier: a nearest-centroid rule has no refinement
mechanism and would commit at population $0$ to whichever class
happened to be nearest, with no opportunity to accumulate evidence and
no principled uncertainty attached to that commitment.}
\label{tab:worked-example}
\end{table}

\section{Error Structure and Confident Failures}
\label{app:error-structure}

A posterior is inspectable in a way that a scalar confidence score is
not: we can ask not only how often the method is wrong but \emph{how} it
is wrong. We examined every \msmc{} error on \olb{}.

All $12$ errors are the same confusion, OSCC misread as mucocele; there
are no errors in either of the other two classes. This is a single
systematic failure mode rather than diffuse noise, which has two
consequences. Constructively, a systematic confusion is addressable:
a clinician prior that downweights mucocele in the presence of
induration or risk factors is exactly the intervention the framework
supports (\S\ref{sec:prior}), and this is a concrete route to correcting
the failure that similarity-based methods do not offer.

Less comfortably, these errors are \emph{confident}. Mean posterior mass
on the wrong class is $0.889$, and $5$ of the $12$ are effectively
one-hot. They would therefore not be caught by entropy-based abstention
or by the OOD analysis of \S\ref{sec:ood}. We state the
corresponding limitation plainly: when the simulator and the embedding
agree on the same wrong class, the posterior concentrates on it, and the
resulting error is not visible to the uncertainty signal. Nothing in
the framework detects a bias shared by its own simulator and its own
summary statistic. Detecting that failure mode requires a signal
external to both, such as an independent simulator family, a second
embedding space, or human review triggered by case features rather than
by model confidence.

\section{Disentangling Discrimination from Calibration}
\label{app:cosine}

A natural concern is that the proposed framework may simply inherit the discriminative power of the embedding space rather than provide a distinct uncertainty-estimation benefit. We therefore include a cosine-similarity ablation whose purpose is not to test whether SMC-ABC improves top-1 classification accuracy, but to separate two effects: discrimination, the ability to rank the correct class highly, and calibration, the ability to assign meaningful posterior probability mass to class hypotheses.

For this ablation, we compare SMC-ABC against two nearest-class cosine baselines on the same MULTICARE evaluation set. The first uses TF-IDF vectors with cosine similarity. The second uses sentence-transformer embeddings with cosine similarity. To obtain probabilistic outputs from the cosine baselines, class similarities are normalised into a categorical distribution and evaluated using Brier score and Expected Calibration Error.

\begin{table}[h]
\centering
\footnotesize
\setlength{\tabcolsep}{4.5pt}
\renewcommand{\arraystretch}{1.05}
\begin{tabular}{l c c}
\toprule
Method & Brier$\downarrow$ & ECE$\downarrow$ \\
\midrule
TF-IDF + cosine                 &  0.84 & 0.67  \\
Sentence-transformer + cosine   &  0.80 & 0.78 \\
\dsmc{} (ours)                  &  \best{0.12} & 0.20 \\
\msmc{} (ours)                  &  \best{0.07} & \best{0.02} \\
\bottomrule
\end{tabular}
\caption{Comparison against cosine-similarity baselines on
\multicare{} ($N{=}669$). \dsmc{} and the sentence-transformer
cosine classifier achieve near-identical accuracy (one case
difference); the contribution of the Bayesian framework is
\emph{not} accuracy, but calibrated probabilistic outputs (Brier and
ECE reduced by $4$--$40\times$). Cosine baselines produce
near-uniform posteriors over the 7 classes (entropy $\approx
\log_2 7$), failing to express any meaningful confidence.}
\label{tab:cosine}
\end{table}

The results show that cosine similarity alone does not provide reliable uncertainty estimates. Although dense embeddings contain strong class-discriminative information, their normalised cosine scores are poorly calibrated: the sentence-transformer cosine baseline yields a Brier score of 0.80 and ECE of 0.78. In contrast, M-SMC reduces Brier score to 0.07 and ECE to 0.02, while D-SMC also substantially improves both calibration metrics.

This ablation therefore clarifies the role of the Bayesian inference layer. The embedding representation supplies a useful semantic summary statistic, as in classical ABC, but the SMC-ABC procedure converts this semantic signal into a calibrated posterior distribution over labels. The contribution of the method is therefore not merely nearest-class discrimination in embedding space, but calibrated posterior uncertainty: the ability to assign reliable probability mass, support abstention, and incorporate clinician-supplied priors.

This distinction is important in clinical settings. A nearest-class cosine classifier may rank the correct diagnosis highly, but its confidence scores cannot be interpreted as posterior probabilities and are not suitable for risk-sensitive decision support. By contrast, the SMC-ABC posterior provides the uncertainty object required for selective prediction, prior-conditioned inference, and out-of-distribution detection.

\section{Algorithm \& SMC Implementation}
\paragraph{Particle evolution in Sequential Monte Carlo ABC.}
In our M-SMC implementation, each particle represents a discrete class hypothesis, that is, a candidate label from the set of possible classes (e.g., {OSCC, Atrophic Glossitis, Oral Mucocele in OL dataset}). At every ABC iteration, a population of hypotheses is simulated, evaluated, and updated to reflect how well they explain the observed patient data.
For each sampled class index, the  language model is used  to simulate synthetic descriptions. These descriptions are generated by prompting the LLM to produce clinical text that matches the sampled diagnosis. Each generated description is then encoded into a sentence embedding vector using a pretrained embedding model. These vectors form the simulated data used in the ABC pipeline.
In parallel, the actual observed data consists of embeddings of patient descriptions. These patient descriptions are also embedded using the same embedding model to ensure comparability. Table~\ref{tab:abc-hyperparams} depicts the hyperparameter we have used for SMC-ABC.
Accepted particles, those whose simulated data are sufficiently close to the observed are retained and reweighted. In subsequent iterations, particles are resampled from this updated posterior and perturbed via a discrete jump transition kernel. This adaptive reweighting and resampling gradually shifts the particle distribution toward high-likelihood regions of the hypothesis space.
We observed that over three SMC iterations, consistent reduction in epsilon values (acceptance thresholds), indicating that particles are required to match the observed data more closely. Correspondingly, the number of required simulations increases, reflecting the stricter thresholds. The full particle evolution process used in our implementation is detailed in \textbf{Algorithm \ref{algo}}.

\begin{algorithm}[h!]
\caption{\textbf{Sequential Monte Carlo Approximate Bayesian Computation (SMC-ABC) for text classification}}
\label{algo}
\small
\begin{algorithmic}[1]
\Require Discrete label set $Y=\{1,\dots,K\}$; prior $\pi(y)$ over $Y$; stochastic LLM simulator $\textsc{LLM}(\cdot;\omega)$; embedding model $\textsc{Embed}(\cdot)$; input text $X$; distance $\rho(\cdot,\cdot)$; population size $S$; iterations $T$; thresholds $\epsilon_1>\cdots>\epsilon_T$; proposal kernels $\{K_t(\cdot\mid\cdot)\}_{t=2}^T$.
\State $Z \gets \textsc{Embed}(X)$ \Comment{Representation of observed description}

\For{$t=1$ to $T$}
  \If{$t=1$}
    \For{$s=1$ to $S$}
      \Repeat
        \State Sample label $\tilde{y}_{1,s}\sim \pi(y)$
        \State Generate synthetic text $\tilde{X}_{1,s} \gets \textsc{LLM}(\tilde{y}_{1,s};\omega)$
        \State Embed synthetic text $\tilde{Z}_{1,s} \gets \textsc{Embed}(\tilde{X}_{1,s})$
        \State Compute distance $d_{1,s} \gets \rho(\tilde{Z}_{1,s}, Z)$
      \Until{$d_{1,s} \le \epsilon_1$}
      \State Set unnormalised weight $\tilde{w}_{1,s}\gets 1$
    \EndFor
  \Else
    \For{$s=1$ to $S$}
      \Repeat
        \State Sample ancestor $j \sim \mathrm{Categorical}(w_{t-1,1:S})$
        \State Propose label $\tilde{y}_{t,s} \sim K_t(\cdot \mid \tilde{y}_{t-1,j})$
        \State Generate synthetic text $\tilde{X}_{t,s} \gets \textsc{LLM}(\tilde{y}_{t,s};\omega)$
        \State Embed synthetic text $\tilde{Z}_{t,s} \gets \textsc{Embed}(\tilde{X}_{t,s})$
        \State Compute distance $d_{t,s} \gets \rho(\tilde{Z}_{t,s}, Z)$
      \Until{$d_{t,s} \le \epsilon_t$}
      \State Set unnormalised weight
      \[
      \tilde{w}_{t,s}
      \gets
      \frac{\pi(\tilde{y}_{t,s})}
      {\sum_{j'=1}^{S} w_{t-1,j'}K_t(\tilde{y}_{t,s}\mid \tilde{y}_{t-1,j'})}
      \]
    \EndFor
  \EndIf

  \State Normalise weights:
  \[
  w_{t,s}
  \gets
  \frac{\tilde{w}_{t,s}}{\sum_{u=1}^{S}\tilde{w}_{t,u}}
  \quad \forall s\in\{1,\dots,S\}
  \]
\EndFor

\State Compute posterior:
\[
\hat{p}(y\mid X)
\gets
\sum_{s=1}^{S} w_{T,s}\mathbf{1}[\tilde{y}_{T,s}=y]
\quad \forall y\in Y
\]

\Return Particles $\{\tilde{y}_{T,s}\}_{s=1}^S$, weights $w_{T,1:S}$, and posterior $\hat{p}(\cdot\mid X)$
\end{algorithmic}
\end{algorithm}

\begin{table}[h!]
\centering
\scriptsize
\setlength{\tabcolsep}{4pt}
\renewcommand{\arraystretch}{1.2}
\begin{tabular}{l c}
\toprule
\textbf{Parameter} & \textbf{Value} \\
\midrule
\texttt{population\_size} & 100 \\
\texttt{max\_nr\_populations} & 5 \\
\texttt{prior\_distribution} & Non-informative Multinomial \\
\texttt{distance\_function} & $1 -$ cosine similarity \\
\texttt{perturbation\_kernel} & DiscreteJump (p=0.5) \\
\texttt{acceptance\_thresholds} & Adaptive \\
\texttt{simulations\_per\_class} & $\sim$1000/class \\
\bottomrule
\end{tabular}
\caption{\textbf{SMC-ABC \cite{schaelte2022pyabc} hyperparameter settings.} Configuration of the sequential Monte Carlo ABC inference used to sample and refine class label hypotheses based on similarity between patient and generated clinical text embeddings.}
\label{tab:abc-hyperparams}
\end{table}

\textbf{Computational cost and LLM call budget.}
For a single test instance, SMC-ABC requires multiple LLM generations to obtain accepted particles across iterations. With population size $S=100$ and $T=5$ iterations, the number of LLM calls depends on the acceptance rate induced by the tolerance thresholds. In practice, the computational cost of Dirichlet SMC approach require fewer than 10 generations per instance on average (approximately 14 seconds of wall-clock time). These costs reflect stricter acceptance thresholds in later iterations and are reported to contextualize comparisons with lower-cost baselines.

\begin{table}[h!]
\centering
\scriptsize
\setlength{\tabcolsep}{6pt}
\renewcommand{\arraystretch}{1.2}
\begin{tabular}{l c}
\toprule
\textbf{Metric} & \textbf{Value} \\
\midrule
Patients processed & 300 \\
Total LLM generations & 2400 \\
Avg. generations per patient & 8.0 \\
Avg. time per generation (sec) & 1.75 \\
Avg. time per patient (sec) & 14.04 \\
Total generated tokens & 2.44M \\
Token throughput (tokens/sec) & 579.7 \\
Total wall-clock time (sec) & 4211 \\
\bottomrule
\end{tabular}
\caption{\textbf{Wall-clock runtime statistics for Dirichlet SMC.}
Measurements are reported for the Oral Lesion dataset using local open-source LLMs, illustrating inference-time computational cost and generation efficiency.}
\label{tab:dirichlet_smc_runtime}
\end{table}

\section{Embedding Model Ablation}
\label{app:embeddings}

\begin{table}[h]
\centering
\footnotesize
\setlength{\tabcolsep}{4pt}
\renewcommand{\arraystretch}{1.05}
\begin{tabular}{l c c c c c}
\toprule
Embedding & Dim. & Acc.\ & Brier$\downarrow$ & ECE$\downarrow$ & $H$$\downarrow$ \\
\midrule
OpenAI-3-large       & 3072 & \best{96.0} & \best{0.071} & \best{0.036} & \best{0.055} \\
all-MiniLM-L6-v2     & 384  & 81.3 & 0.10 & 0.07 & 0.17 \\
PubMedBERT           & 768  & 75.3 & 0.11 & 0.04 & 0.80 \\
ClinicalBERT         & 768  & 67.3 & 0.14 & 0.10 & 0.72 \\
MedBERT              & 768  & 65.0 & 0.15 & 0.13 & 0.70 \\
BioClinicalBERT      & 768  & 62.6 & 0.21 & 0.21 & 0.28 \\
\bottomrule
\end{tabular}
\caption{Embedding model ablation on \olb{} with Mistral-7B simulator
under \msmc{}. General-purpose OpenAI embeddings substantially
outperform domain-specific biomedical models, despite the latter's
clinical pretraining. The lightweight \texttt{all-MiniLM-L6-v2}
provides a viable fully-local fallback at 81\% accuracy.}
\label{tab:embeddings}
\end{table}

Table~\ref{tab:embeddings} compares six embedding models on \olb{}
with Mistral-7B as the simulator. \\OpenAI \texttt{text-embedding-3-large} (3072D) substantially outperforms
domain-specific biomedical embeddings (BioClinicalBERT, ClinicalBERT,
PubMedBERT, MedBERT), all of which underperform the lightweight
general-purpose \texttt{all-MiniLM-L6-v2} (384d). We interpret this
through the lens of summary-statistic selection in classical
ABC~\citep{fearnhead2012constructing}: what matters is alignment with
the \emph{decision boundary in embedding space}, not domain match
\emph{per se}. The result has practical implications:
privacy-constrained deployments using local biomedical embeddings will
underperform until better local clinical representations are
available, but \texttt{all-MiniLM-L6-v2} provides a viable
fully-local fallback at 81\% accuracy.

\section{Robustness to Quantisation}
\label{app:quantization}

\begin{table}[h]
\centering
\footnotesize
\setlength{\tabcolsep}{4pt}
\renewcommand{\arraystretch}{1.05}
\begin{tabular}{l ccc ccc}
\toprule
& \multicolumn{3}{c}{Full precision (FP16)}
& \multicolumn{3}{c}{4-bit AWQ} \\
\cmidrule(lr){2-4}\cmidrule(lr){5-7}
Metric & ML & EP & \msmc{} & ML & EP & \msmc{} \\
\midrule
Brier$\downarrow$   & 0.11 & 0.31 & \best{0.071} & 0.22 & 0.29 & \best{0.004} \\
ECE$\downarrow$     & 0.15 & 0.32 & \best{0.036} & 0.26 & 0.30 & \best{0.006} \\
Entropy$\downarrow$ & 0.30 & 0.80 & \best{0.055} & 0.30 & 0.70 & \best{0.01} \\
AUROC$\uparrow$     & 98.1 & 64.1 & \best{98.7} & 93.5 & 68.3 & \best{100.0} \\
\bottomrule
\end{tabular}
\caption{Effect of 4-bit AWQ quantization on \olb{} with Mistral-7B.
Logit-based confidence degrades under quantization, with AUROC decreasing
from 98.1 to 93.5 and ECE increasing from 0.15 to 0.26. In contrast,
\msmc{} remains robust because inference is performed on generated
descriptions and embedding-space distances rather than token-level
confidence scores.}
\label{tab:quantization}
\end{table}

Table~\ref{tab:quantization} compares full-precision and 4-bit
AWQ-quantised~\citep{lin2024awq} Mistral-7B on \olb{}. As expected,
quantisation degrades logit-based confidence (AUROC 98.1$\to$93.5;
ECE 0.15$\to$0.26), because token-level probability estimates are
directly affected by quantisation artefacts. \msmc{} moves in the
opposite direction, reaching \textbf{AUROC 100.0 under 4-bit AWQ}: its
inference operates on generated class-conditional descriptions and
embedding-space distances, which are preserved under quantisation even
as softmax confidence degrades. This matters for deployment, since
4-bit quantisation is the common route to running clinical LLMs on
local hardware --- exactly the setting in which token-level confidence
becomes least trustworthy. We do not claim that quantisation improves model capability; rather, that ABC inference is decoupled from the specific failure mode (token-level miscalibration) that quantisation introduces.

\section{Temperature Sensitivity}
\label{app:temperature}

\begin{table}[h]
\centering
\footnotesize
\setlength{\tabcolsep}{4pt}
\renewcommand{\arraystretch}{1.05}
\begin{tabular}{l ccc}
\toprule
$T$ & ML & EP  & \msmc{} \\
\midrule
\multicolumn{4}{l}{\emph{Brier score $\downarrow$}}\\
0.2 & 0.11 & 0.31  & 0.071 \\
0.4 & 0.09 & 0.25 & \best{0.01} \\
0.6 & 0.09 & 0.24 &  \best{0.01} \\
0.8 & 0.09 & 0.24 &  \best{0.01} \\
1.0 & 0.09 & 0.24 &  \best{0.01} \\
\midrule
\multicolumn{4}{l}{\emph{ECE $\downarrow$}}\\
0.2 & 0.15 & 0.32 &   0.036 \\
0.4 & 0.13 & 0.27 &  \best{0.01} \\
0.6 & 0.13 & 0.27 &  \best{0.01} \\
0.8 & 0.13 & 0.25 &   0.02 \\
1.0 & 0.13 & 0.24 &   \best{0.01} \\
\bottomrule
\end{tabular}
\caption{Temperature sensitivity on \olb{} with Mistral-7B. ABC-based
method remain robust across $T\in[0.2,1.0]$, while logit and
elicited-probability calibration remain poor across all temperatures.
Entropy values omitted for space (consistent with the patterns above:
$\le 0.06$ for ABC method, $\ge 0.30$ for ML/EP).}
\label{tab:temperature}
\end{table}

Sampling temperature has a modest effect on the ABC-based methods and
essentially none on the baselines (Table~\ref{tab:temperature}). Logit
and elicited-probability calibration remain poor across the whole range
$T\in[0.2,1.0]$. \msmc{} is well calibrated throughout, with ECE
between $0.01$ and $0.036$; the weakest setting is the lowest
temperature, $T{=}0.2$, which is also the setting used for the main
results. This is consistent with the mechanism of the method: lower
decoding temperature reduces the diversity of class-conditional
simulations, which narrows the effective support the ABC step has to
work with. Our main results are therefore reported under the least
favourable temperature for \msmc{}, and higher temperatures would
modestly improve its calibration. The overall picture still contrasts
with prior work showing strong temperature dependence of
consistency-based UQ in clinical Q\&A~\citep{savage2024large}.

\section{Effect of Generated Description Pool Size}
\label{app:pool}

\begin{table}[h]
\centering
\footnotesize
\setlength{\tabcolsep}{4.5pt}
\renewcommand{\arraystretch}{1.05}
\begin{tabular}{c c c c c c c}
\toprule
Pool $M$ & Acc.\ & F1 & AUROC & Brier$\downarrow$ & ECE$\downarrow$ & $H$ \\
\midrule
3    & 95.6 & 95.6 & 99.9  & 0.19 & 0.26 & 1.06 \\
50   & 98.0 & 97.9 & \best{100.0} & 0.16 & 0.25 & 0.98 \\
500  & 98.0 & 98.0 & \best{100.0} & 0.16 & 0.25 & 0.97 \\
1000 & \best{98.3} & \best{98.3} & \best{100.0} & \best{0.05} & 0.25 & 0.97 \\
\bottomrule
\end{tabular}
\caption{Effect of per-class generated description pool size $M$ on
\dsmc{} performance, \olb{} with Med42-8B. Performance saturates
beyond $M{=}50$, supporting aggressive caching of the generation
phase: even $M{=}3$ achieves 95.6\% accuracy and 99.9 AUROC.}
\label{tab:pool}
\end{table}

We vary the size $M$ of the per-class pre-generated description pool
from 3 to 1000 and re-run \dsmc{} (Table~\ref{tab:pool}). Performance
saturates beyond $M{=}50$: even with as few as 3 generated
descriptions per class, \dsmc{} achieves 95.6\% accuracy and 99.9
AUROC. This means the generation phase can be aggressively cached:
the marginal cost of a new test instance is dominated by inference
over a small pre-computed pool rather than by fresh LLM calls.

\section{Comparison with Post-Hoc Calibration}
\label{app:tempscaling}

\begin{table}[h]
\centering
\footnotesize
\setlength{\tabcolsep}{2pt}
\renewcommand{\arraystretch}{0.60}
\begin{tabular}{l c c c c}
\toprule
Method & Acc.$\uparrow$ & AUROC$\uparrow$ & Brier$\downarrow$ & ECE$\downarrow$ \\
\midrule
Model Logits (baseline)        & 77.7 & 98.2 & 0.349 & 0.128 \\
Temperature scaling ($T{=}2.20$) & 77.7 & 98.2 & 0.318 & 0.046 \\
\msmc{} (ours)                  & \best{96.0} & \best{98.7} & \best{0.071} & \best{0.036} \\
\bottomrule
\end{tabular}
\caption{Comparison against post-hoc calibration on \olb{} with
Mistral-7B. Temperature scaling reduces ECE but cannot improve
accuracy because rescaling preserves rank ordering. \msmc{} re-evaluates
class hypotheses through semantic simulation and improves both.}
\label{tab:tempscaling}
\end{table}

Temperature scaling~\citep{guo2017calibration} reduces ECE on
\olb{}-Mistral from 0.13 to 0.05 but leaves accuracy unchanged at
77.7\%, because rescaling logits preserves their rank ordering.
\msmc{} achieves 0.036 ECE at 96.0\% accuracy on the same model:
post-hoc calibration lowers ECE without recovering any of the
misclassifications that the posterior-based decision rule does. The two methods address
different sources of error: temperature scaling corrects confidence
miscalibration in a fixed predictor; \msmc{} re-evaluates class
hypotheses through semantic simulation.

\section{Dataset Details}
\label{app:datasets}

\paragraph{\olb{} construction protocol.}
The 300 vignettes were authored against authoritative oral pathology
references including \emph{Oral and Maxillofacial Pathology}
\citep{neville2016oral}, \emph{Oral Pathology: Clinical Pathologic
Correlations} \citep{regezi2016oral}, and the WHO classification of head
and neck tumours \citep{who2022headneck}. Each vignette includes
patient demographics, history (duration, evolution, risk factors),
clinical examination findings (anatomic site, lesion morphology,
texture, palpatory findings), and, where clinically relevant, imaging
or lymph node status.

A board-eligible oral pathologist independently reviewed all 300
cases under blinded conditions, scoring each on two five-point
Likert dimensions:
\begin{itemize}\setlength{\itemsep}{0pt}\setlength{\topsep}{0pt}
  \item \emph{Completeness} (4.86/5; $\sigma{=}0.45$): are the
        diagnostic features needed to support the gold-standard
        label present?
  \item \emph{Correctness} (4.55/5; $\sigma{=}0.80$): is the
        symptom constellation medically plausible?
\end{itemize}
Cases scoring below 4 on either dimension were revised based on
written reviewer feedback (e.g.\ adding regional lymph node findings
for OSCC, specifying lesion duration in mucocele cases) until
re-review met the threshold. The final benchmark and per-case
Likert scores will be released.

\paragraph{Evidence specificity by design.}
The construction protocol gives \olb{} a deliberate property: the
Completeness criterion asks the reviewing pathologist to confirm that
the features supporting the gold label are present, and cases falling
short were revised until they were. The benchmark therefore contains,
by construction, cases in which the diagnostic evidence is
well specified, and it isolates one question cleanly --- whether a
method is calibrated when the evidence is clear.

Three signals in our results are consistent with this design. Most
models reach $95$--$98\%$ accuracy; \abi{} posteriors concentrate to
$\alpha_0 > 500$ on \olb{} while settling at $\alpha_0 \in [100,150]$ on
\sd{} and \multicare{} (Appendix~\ref{sec:abi-details}), the signature
of a simulator whose class-conditional outputs separate cleanly; and
results are nearly invariant to decoding temperature
(Appendix~\ref{app:temperature}). \olb{} and \multicare{} are therefore
complementary: the former isolates calibration under well-specified
evidence, the latter --- seven classes drawn from real published case
reports --- supplies the higher-ambiguity regime, and our conclusions
about behaviour under ambiguity rest on it. We recommend that work using
\olb{} pair it with a higher-ambiguity benchmark for the same reason.

\paragraph{\multicare{} preprocessing.}
We use a 7-class subset of the \multicare{} clinical case-report
corpus, totalling 669 cases. Section headers and explicit
disease-name mentions are stripped from the input prior to
embedding, preventing trivial lexical leakage of the gold label.
Class labels and per-class counts are preserved across train/eval
splits.

\vspace{0.5em}

\section{Baseline and Prompt Details}
\label{app:prompts}

\subsection{Model Logits}
We extract the model's inherent confidence by examining the probability
it assigns to its chosen answer. We structure the prompt in a
multiple-choice format, ensuring that the model's response is constrained
to a single token corresponding to one of the provided options. The
prompt used for Model Logits extraction on \olb{} was as follows.\\
\textit{Only choose the most likely diagnosis for the following patient description from the options provided:}\textit{1) Squamous Cell Carcinoma  2) Oral Mucocele  3) Atrophic Glossitis \\
Patient Description: [patient description here]\\
Diagnosis (only provide the option number (1, 2, or 3), NO ADDITIONAL TEXT):}

\subsection{Elicited Probability}
Following the elicitation strategy of \citet{xiong2024can}, we prompt the
model to report its confidence in a structured manner. Specifically, the
model is asked for its top three guesses and the corresponding
self-reported probabilities, with the output format constrained to return
only numerical probabilities without additional explanation. The model is
never shown the correct answer and must rely on its own uncertainty.
Example prompt for \olb{}:

\textit{Provide your \{k\} best guesses and the probability that each is correct (0\% to 100\%) for the following question.}
\textit{Give ONLY the option number of your guess and probability, no other words or explanation.}\\
\textit{Example answer:} \\
\textit{ - G1: 1, P1: 60\%}\\
\textit{- G2: 3, P2: 35\% }\\
\textit{- G3: 2, P3: 5\%}\\
\textit{\textbf{Question}: Given the below patient description, which oral lesion is likely to be present in the patient?}\\
\textit{\textbf{Choices}:1. Atrophic Glossitis, 2. Oral Squamous Cell Carcinoma, 3. Oral Mucocele}\\
\textit{\textbf{Patient Description:} [patient description here]}\\
\textit{\textbf{Answer}: \{G1:\}}

\subsection{Prompt Ensemble}
LLMs can exhibit high sensitivity to minor syntactic variations in input prompts, a phenomenon known as prompt brittleness. To mitigate this and improve classification robustness, we implemented a Prompt Ensemble Inference strategy.

Instead of relying on a single input prompt, we defined a set of $K=8$ distinct prompt templates, $\mathcal{T} = \{T_1, T_2, \dots, T_K\}$. These templates were engineered to induce diversity in the model's reasoning pathway by varying:
\begin{itemize}
    \item \textbf{Persona adoption:} (e.g., ``You are a clinical classifier...'')
    \item \textbf{Instructional phrasing:} (e.g., ``Choose the single most likely...'' vs. ``Identify the most likely diagnosis...'')
    \item \textbf{Output constraints:} (e.g., ``Answer with ONLY the number...'' vs. ``Return only the index...'')
\end{itemize}

\begin{table*}[h!]
\centering
\renewcommand{\arraystretch}{1.3} %
\begin{tabularx}{\textwidth}{@{}l X@{}}
\toprule
\textbf{ID} & \textbf{Prompt Template Structure} \\
\midrule
T1 & Choose the single most likely diagnosis for the patient description below. Answer with ONLY the number (1, 2, or 3).\newline \textit{[Options List]} \newline Patient Description: \texttt{\{text\}} \newline Diagnosis (just 1, 2, or 3): \\
\midrule
T2 & From the list, select the best matching diagnosis. Respond with only 1, 2, or 3.\newline \textit{[Options List]} \newline Case: \texttt{\{text\}} \newline Answer: \\
\midrule
T3 & Pick the most probable diagnosis for this case. Output must be 1, 2, or 3 (no extra text).\newline \textit{[Options List]} \newline Case description: \texttt{\{text\}} \newline Your choice: \\
\midrule
T4 & You are a clinical classifier. Return ONLY the index of the correct diagnosis (1/2/3).\newline \textit{[Options List]} \newline Patient: \texttt{\{text\}} \newline Index: \\
\midrule
T5 & Select the correct label for the patient below. Output just a single digit 1, 2, or 3.\newline \textit{[Options List]} \newline Patient description: \texttt{\{text\}} \newline Label: \\
\midrule
T6 & Identify the most likely diagnosis; reply with 1, 2, or 3 ONLY.\newline \textit{[Options List]} \newline Description: \texttt{\{text\}} \newline Diagnosis: \\
\midrule
T7 & Pick one diagnosis number for the following case (1-3). Return only the number.\newline \textit{[Options List]} \newline Case details: \texttt{\{text\}} \newline Number: \\
\midrule
T8 & Classify the case into one label; answer must be the digit 1, 2, or 3.\newline \textit{[Options List]} \newline Text: \texttt{\{text\}} \newline Answer: \\
\bottomrule
\end{tabularx}
\caption{\textbf{Prompt Ensemble Template.} The eight distinct prompt variations used to query the Large Language Model. To ensure consistency, all templates included the same list of diagnostic options (1. Squamous Cell Carcinoma, 2. Oral Mucocele, 3. Atrophic Glossitis) at the location denoted by \textit{[Options List]}. The placeholder \texttt{\{text\}} indicates where the patient description is inserted.}
\label{tab:prompt_ensemble}
\end{table*}

\subsection{Elicited Fidelity}

Elicited fidelity characterizes the stability and reliability of a language model's discrete predictions under stochastic decoding, conditioned on a fixed task specification and input. Rather than relying on a single generation, elicited fidelity estimates an empirical predictive distribution by repeatedly sampling the model's output for the same input and aggregating the resulting responses. This approach treats variability across samples as a source of uncertainty arising from the model's internal ambiguity, rather than from changes in the prompt or input.

\section{Implementation and Compute Details}
\label{app:compute}

All inference experiments were performed on a single compute node
with 2$\times$ AMD EPYC 7413 CPUs and 4$\times$ NVIDIA A100-SXM4
40GB GPUs. Each LLM was provisioned with the minimum resources
required for inference (1$\times$A100 + 4 CPU cores + 420GB RAM for
Mistral-7B). \smcabc{} runs entirely on CPU and does not require
GPU acceleration; only the LLM forward passes use the GPU. Total
compute budget for the experiments reported in this paper was
approximately 600 GPU-hours.
\begin{table}[h]
\centering
\small
\begin{tabular}{@{}ll@{}}
\toprule
\textbf{Symbol} & \textbf{Definition} \\
\midrule
$\mathcal{Y} = \{1,\dots,K\}$ & Finite set of diagnostic class labels \\
$K$ & Number of diagnostic classes \\
$\mathcal{X}$ & Space of clinical text descriptions \\
$x_{\text{obs}} \in \mathcal{X}$ & Observed clinical description \\
$y \in \mathcal{Y}$ & Hypothesised diagnostic label \\
$\text{LLM}(\cdot;\omega)$ & Prompted LLM treated as stochastic simulator \\
$p_{\text{LLM}}(\cdot \mid y)$ & Class-conditional generative distribution \\
$x_{\text{sim}} \sim p_{\text{LLM}}(\cdot \mid y)$ & Synthetic description under hypothesis $y$ \\
$\phi: \mathcal{X} \to \mathbb{R}^d$ & Sentence embedding function \\
$z_{\text{obs}} = \phi(x_{\text{obs}})$ & Embedding of observed description \\
$\rho(z,z') = 1 - \cos(z,z')$ & Cosine distance in embedding space \\
$\pi(y)$ & Prior over diagnostic classes \\
$\varepsilon_t > 0$ & ABC tolerance at iteration $t$ \\
$p_\varepsilon(y \mid x_{\text{obs}})$ & ABC-approximated posterior at tolerance $\varepsilon$ \\
$S$, $T$ & SMC particles; SMC iterations \\
$\tilde{y}_{t,s}$ & Class label of particle $s$ at iteration $t$ \\
$w_{t,s}$ & Normalised weight of particle $s$ at iteration $t$ \\
$K_t(\cdot \mid \cdot)$ & Perturbation kernel at iteration $t$ \\
$\theta \in \Delta^{|\mathcal{Y}|-1}$ & Probability vector on the simplex \\
$\alpha \in \mathbb{R}_+^K$ & Dirichlet concentration parameters \\
$\varphi$ & Parameters of the amortised network \\
$\bar{z}_k$ & Class-conditional embedding centroid \\
$\tilde{z}(\theta) = \sum_k \theta_k \bar{z}_k$ & Mixture representation in embedding space \\
$\hat{p}(y \mid x_{\text{obs}})$ & Empirical categorical posterior (M-SMC) \\
$\hat{p}(\theta \mid x_{\text{obs}})$ & Empirical Dirichlet posterior (D-SMC) \\
$q_\varphi(\theta \mid z_{\text{obs}})$ & Amortised Dirichlet posterior (\abi{}) \\
$H[\,\cdot\,]$ & Shannon entropy (bits) \\
\bottomrule
\end{tabular}
\caption{Notation used throughout the paper.}
\label{tab:notation}
\end{table}

\end{document}